\documentclass[11pt]{article}
\usepackage{amsmath}

\usepackage[final]{acl}

\usepackage{times}
\usepackage{latexsym}
\usepackage{subcaption}
\usepackage{amssymb}
\usepackage[most]{tcolorbox}
\tcbuselibrary{breakable}
\usepackage{xurl}
\usepackage{xcolor}
\usepackage[T1]{fontenc}

\usepackage[utf8]{inputenc}
\usepackage{booktabs}

\usepackage{microtype}
\usepackage{multirow}

\usepackage{inconsolata}

\usepackage{graphicx}

\usepackage[most]{tcolorbox}

\usepackage{tcolorbox}
\tcbuselibrary{listings, breakable}

\newtcolorbox{promptbox}[2][]{enhanced,
  breakable,
  colback=gray!5,
  colframe=black!50,
  fonttitle=\bfseries,
  title={#2},
  #1}
  
\usepackage{pgfplots}
\pgfplotsset{compat=1.18}

\title{\textit{When Reviews Disagree:} Fine-Grained Contradiction Analysis in Scientific Peer Reviews}

\author{
Sandeep Kumar$\dagger$, 
Yash Kamdar$\dagger$, 
Abid Hossain$\dagger$, 
Bharti Kumari$\ddagger$, 
Tanik Saikh$\ddagger$,  
Asif Ekbal$\dagger$ 
\\
$\dagger$Department of Computer Science and Engineering, Indian Institute of Technology Patna, India \\
$\ddagger$School of Computer Engineering, KIIT Deemed to be University, Bhubaneswar, India\\
\texttt{\{sandeep\_2121cs29, 2201ai45\_yash, abid\_2311ai22, asif\}@iitp.ac.in} \\
\texttt{\{22052975, tanik.saikhfcs\}@kiit.ac.in}
}

\begin{document}
\maketitle
\begin{abstract}
Scientific peer reviews frequently contain conflicting expert judgments, and the increasing scale of conference submissions makes it challenging for Area Chairs and editors to reliably identify and interpret such disagreements. Existing approaches typically frame reviewer disagreement as binary contradiction detection over isolated sentence pairs, abstracting away the review-level context and obscuring differences in the severity of evaluative conflict. In this work, we introduce a fine-grained formulation of reviewer contradiction analysis that operates over full peer reviews by explicitly identifying contradiction evidence spans and assigning graded disagreement intensity scores. To support this task, we present \textbf{RevCI}, an expert-annotated benchmark of peer-review pairs with evidence-level contradiction annotations with graded intensity labels. We further propose \textbf{IMPACT}, a structured multi-agent framework that integrates aspect-conditioned evidence extraction, deliberative reasoning, and adjudication to model reviewer contradictions and their intensity. To support efficient deployment, we distill IMPACT into \textbf{TIDE}, a small language model that predicts contradiction evidence and intensity in a single forward pass. Experimental results show that IMPACT substantially outperforms strong single-agent and generic multi-agent baselines in both evidence identification and intensity agreement, while TIDE achieves competitive performance at significantly lower inference cost. We make our code and  dataset publicly available\footnote{\url{https://github.com/sandeep82945/Contradiction-Intensity.git}}.

\end{abstract}

\section{Introduction}





Scientific peer review serves as the cornerstone of academic publishing, ensuring the quality and integrity of disseminated research \cite{DBLP:journals/arist/Bornmann11}. Despite its status as the \textit{de facto} standard for validating scholarly research, the traditional peer-review process has come under increasing scrutiny due to perceived systemic fragilities. These concerns range from a lack of transparency \cite{980f5632f0664f9681dea65f7a810bff, fbdf6c99376b42789f9a8e5305d8afed} and inherent biases \cite{DBLP:journals/pacmhci/StelmakhSSD21, DBLP:conf/nips/StelmakhSS19} to the fundamental arbitrariness \cite{DBLP:journals/scientometrics/BrezisB20} and inconsistency of reviewer evaluations \cite{DBLP:journals/jmlr/ShahTMGL18, DBLP:journals/cacm/LangfordG15}. Furthermore, peer review is often critiqued as an ill-defined task \cite{rogers-augenstein-2020-improve}, frequently failing to identify high-impact or influential contributions in their early stages \cite{DBLP:journals/cacm/FreyneCSC10}.

In recent years, the computational linguistics (CL) community has experienced an unprecedented surge in submission volumes, placing the peer review ecosystem under severe strain \cite{peer_review_stress, 10.1093/biosci/bix034}. Large-scale venues such as EMNLP 2025 now coordinate review cycles involving over 13,000 reviewers, exacerbating logistical challenges and contributing to declining review quality, including instances of “highly irresponsible” reviewing behavior \cite{emnlp2025chairs}. At the same time, the integrity of peer review is increasingly threatened by the unauthorized use of Large Language Models (LLMs) to generate reviews, which often produce superficial or formulaic feedback lacking substantive critical engagement \cite{liang2024monitoring, Mishra_2025}. Beyond these operational pressures, the peer review process is inherently subjective, frequently giving rise to reviewer disagreement, where experts offer conflicting assessments of the same manuscript \cite{rogers-augenstein-2020-improve}. For Area Chairs and editors, reconciling these contradictions is the most labor-intensive aspect of the decision-making process. While some level of disagreement is expected in cutting-edge research, unresolved or unexplained conflicts can lead to inconsistent outcomes and a perceived lack of fairness in the evaluation system \cite{10.1145/3209978.3210056}. Prior work has begun to explore computational methods for identifying reviewer disagreement in scientific peer review. Notably, ContraSciView~\cite{kumar-etal-2023-reviewers} formulates disagreement as a binary contradiction detection task over isolated sentence pairs. While effective for identifying explicit sentence-level conflicts, this formulation abstracts away review-level discourse and collapses graded differences in evaluative judgments into coarse binary labels, limiting its usefulness for Area Chairs who must reason about both the presence and the severity of conflicting reviews.

In this study, we introduce \textbf{IMPACT} (Intensity-based Multi-Agent Contradiction estimation), a structured multi-agent framework for fine-grained analysis of reviewer disagreement. Unlike prior approaches that reduce contradictions to binary labels over isolated sentence pairs, \textbf{IMPACT} operates over full review contexts to jointly identify aspect-conditioned contradiction evidence spans, generate natural-language explanations, and assign graded disagreement intensity scores. The framework is driven by a deliberative reasoning protocol in which multiple agents explicitly debate conflicting assessments, helping to mitigate individual biases and surface nuanced disagreements. To support this task, we present \textbf{RevCI}, an expert-annotated dataset of peer-review pairs with evidence-level contradiction annotations and graded intensity labels. Finally, to support efficient deployment, we introduce \textbf{TIDE}, a small language model (SLM) distilled from \textbf{IMPACT}'s deliberative reasoning traces. Experimental results show that \textbf{IMPACT} substantially outperforms existing baselines in contradiction detection, while \textbf{TIDE} achieves performance comparable to much larger models at a fraction of the inference cost.

We summarize our contributions as follows:
\begin{itemize}
    \item We introduce a new task for analyzing reviewer disagreement that moves beyond binary contradiction detection by jointly identifying explicit contradiction evidence and assigning graded disagreement intensity scores in scientific peer reviews.
    
    \item We present \textbf{RevCI}, an expert-annotated benchmark of peer reviews that enables fine-grained analysis of reviewer disagreement through evidence-level contradiction annotations, graded intensity labels, and human-written explanation statements.
    
    \item We propose \textbf{IMPACT}, a structured multi-agent framework that integrates aspect-conditioned evidence extraction, deliberative reasoning, and adjudication to detect contradiction evidence and estimate disagreement intensity from full review contexts, substantially outperforming strong single-agent and generic multi-agent baselines.
    
    \item To support efficient deployment, we introduce \textbf{TIDE}, a small language model distilled from \textbf{IMPACT} that predicts contradiction evidence and intensity in a single forward pass, achieving competitive agreement with human annotations at significantly lower inference cost.
\end{itemize}

\section{Related Work}
Reviewer disagreement has been studied in the context of peer review through analyses of score variance \cite{bornmann2010usefulness}, sentiment divergence \cite{kang-etal-2018-dataset}, and reviewer bias \cite{DBLP:conf/nips/StelmakhSS19,DBLP:journals/jmlr/ShahTMGL18}. While these studies characterize disagreement at an aggregate level, they do not explicitly model textual contradictions between reviewer comments. Contradiction detection is commonly studied under the Natural Language Inference (NLI) framework \cite{DBLP:conf/emnlp/BowmanAPM15,DBLP:conf/naacl/WilliamsNB18}, with transformer-based models achieving strong performance on benchmark datasets \citep{DBLP:conf/naacl/DevlinCLT19,DBLP:journals/corr/abs-1907-11692}. However, prior work has shown that such models struggle with pragmatic and domain-specific contradictions in expert-written text \citep{DBLP:conf/acl/NieWDBWK20,DBLP:conf/acl/GururanganMSLBD20}. Peer reviews pose additional challenges due to hedging, technical assumptions, and discourse-level reasoning, which are difficult to capture using sentence-pair classification alone. 


The work most closely related to ours is ContraSciView \citep{kumar-etal-2023-reviewers}, which frames reviewer disagreement as a contradiction detection problem. ContraSciView formulates the task as binary classification over isolated review sentence pairs using encoder-based models, such as BERT \cite{DBLP:conf/naacl/DevlinCLT19}. While effective for detecting explicit sentence-level conflicts, this formulation does not capture discourse-level disagreement expressed across full reviews, including hedged, comparative, and assumption-dependent judgments.

\section{RevCI Dataset}
We introduce RevCI (Review Contradiction Intensity), a dataset of expert-annotated peer-review pairs with human-written contradiction summaries, contradiction evidence pairs, and graded intensity labels.

\subsection{Dataset Collection and Re-Annotation}
\label{sec:data_collection}

To annotate RevCI, we reuse the peer-review dataset from ContraSciView, derived from the ASAP-Review corpus \cite{DBLP:journals/corr/abs-2102-00176}. The dataset contains reviews from 8,582 papers across ICLR (2017--2020) and NeurIPS (2016--2019). While the underlying data source remains unchanged, we conduct a new round of human annotation to support a more fine-grained analysis of reviewer disagreement. In contrast to prior work that focused on binary contradiction detection, our annotations capture evidence-grounded contradiction statements and graded intensity levels.

\subsection{Review Pair Construction and LLM-Based Filtering} \label{sec:pair_construction}

We define a \textit{review} as the set of comments authored by a single reviewer. For a paper with $n$ reviews, we construct all unordered pairs of reviews, resulting in $\binom{n}{2}$ review pairs per paper and approximately 28K review pairs across the dataset. Each review pair contains multiple candidate comment-level pairs formed by pairing individual comments from the two reviews.

Explicit contradictions between peer reviews are relatively rare, making uniform sampling inefficient for annotation. To address this, we apply an instruction-following LLM\footnote{We used GPT-4o mini.} as a screening model prior to human annotation. Given a review pair, the model predicts whether the pair contains a contradiction, and only pairs flagged as potentially contradictory are forwarded for annotation.

Due to space limitations, we discuss the \textbf{annotation guidelines, annotation process, and annotator compensation} in detail in Appendix~\ref{Appendix: annotation_details}.

\subsection{Final Dataset}

The final dataset comprises 800 review pairs\footnote{RevCI contains 800 review pairs: 352 contain at least one contradiction, while the remaining 448 contain none and serve as negative instances for contradiction detection and FPR computation.}. While RevCI is modest in size, its reliance on expert-written peer reviews and evidence-level, graded annotations makes large-scale collection costly, and similar dataset scales are standard for tasks requiring detailed expert judgment, as seen in fine-grained evaluation benchmarks such as SummEval~\cite{fabbri2020summeval} and Qasper~\cite{DBLP:journals/corr/abs-2105-03011}. Appendix, Figures~\ref{fig:aspect_dist_training} and~\ref{fig:intensity_dist_training} summarize the annotation statistics for the contradiction-bearing subset.\footnote{Figures~\ref{fig:aspect_dist_training} and~\ref{fig:intensity_dist_training} are computed over the contradiction-bearing subset (352 pairs).}


\section{Methodology}

\begin{figure*}[ht]
\centering
  \includegraphics[width=2.0\columnwidth]{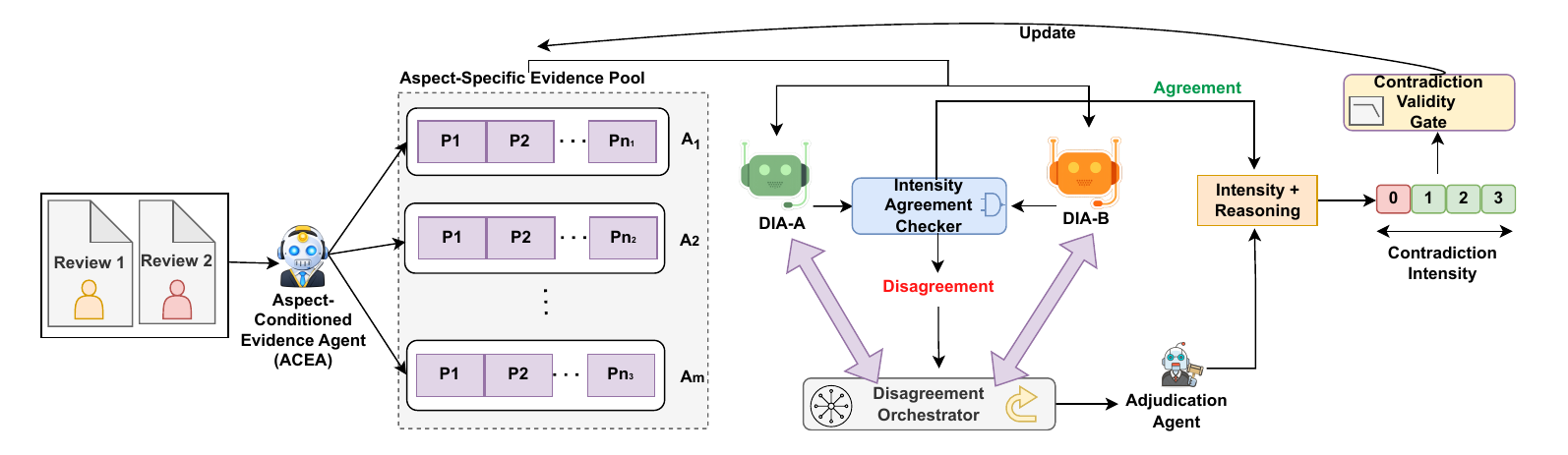}
  \caption{Overview of \textbf{IMPACT}, an Intensity-based Multi-Agent Contradiction estimation framework. The framework integrates aspect-conditioned evidence extraction with structured multi-agent disagreement to estimate contradiction intensity.}
  \label{fig: multiagent_framework}
\end{figure*}

\subsection{Problem Formulation}


Given two full peer reviews $r_i$ and $r_j$, where i and j index distinct reviews of the same manuscript, our goal is to identify and characterize fine-grained contradictions between them. Unlike prior approaches that operate on isolated sentence pairs or assume pre-segmented review comments, we process full reviews end-to-end and generate a set of contradiction evidence pairs $E=\{(e_1^{(t)}, e_2^{(t)})\}_{t=1}^{m}$. Here, m denotes the total number of contradiction pairs, and each pair index t corresponds to a distinct disagreement instance between the two reviews. Each pair consists of atomic evidence statements grounded in $r_i$ and $r_j$ that express mutually incompatible judgments about the same underlying issue. Each evidence pair is associated with an aspect category $a_t \in \mathcal{A}$ (e.g., novelty, clarity, soundness), a discrete contradiction intensity score $\alpha_t \in \{1,2,3\}$, and a natural-language rationale $\rho_t$ explaining the source and severity of the disagreement. The final task output is the collection $\{(e_1^{(t)}, e_2^{(t)}, a_t, \alpha_t, \rho_t)\}_{t=1}^{m}$, providing an aspect-aware, graded representation of reviewer disagreement.

\subsection{Overall}
Our methodology consists of two complementary components with distinct inference-time trade-offs. \textbf{IMPACT} is a standalone multi-agent framework that estimates contradiction intensity at inference time via aspect-conditioned evidence extraction, structured agent disagreement, and adjudication. Although this multi-agent deliberation incurs higher inference latency due to iterative agent interactions, it yields explicit, evidence-grounded reasoning traces. To enable efficient deployment, we further introduce \textbf{TIDE}, a Small Language Model (SLM) trained to directly predict contradiction evidence, intensity labels, and associated reasoning distilled from \textbf{IMPACT}, resulting in substantially faster inference.

\subsection{IMPACT: Intensity-based Multi-Agent Contradiction Estimation}
We propose \textsc{IMPACT} (\textit{Intensity-based Multi-Agent Contradiction estimation}), a multi-agent framework for estimating contradiction intensity by coordinating aspect-conditioned evidence extraction with structured agent disagreement and adjudication. IMPACT consists of an Aspect-Conditioned Evidence Agent (ACEA), two Deliberative Intensity Agents (DIA-A/B) coordinated by an Intensity Agreement Checker, a Disagreement Orchestrator, an Adjudication Agent for resolution, and a final Contradiction Validity Gate.

\paragraph{Aspect-Conditioned Evidence Agent (ACEA):}
ACEA operates on a single pair of reviews and serves as the high-recall evidence retrieval
component of IMPACT. Given a review pair $(r_i, r_j)$ and a predefined set of aspects $\mathcal{A}=\{a_1,\dots,a_M\}$, ACEA conditions on each aspect $a_m$ to identify candidate pairs that exhibit potential conflicting viewpoints. For each aspect $a_m$, ACEA produces paired evidence spans:
\begin{equation}
\small
\mathcal{E}_{a_m}^{(i,j)} =
f_{\textsc{ACEA}}(r_i, r_j, a_m)
\subseteq \mathrm{Spans}(r_i) \times \mathrm{Spans}(r_j),
\end{equation}
where $f_{\textsc{ACEA}}$ denotes an LLM-based extraction function configured to retrieve
pairs exhibiting potential semantic divergence.
This high-recall filtering ensures that the system retains subtle contradictions while discarding clear agreements, providing a unified candidate set for intensity assessment. Across review pairs, extracted evidence is accumulated into an \emph{Aspect-Specific Evidence Pool}
$\mathcal{E}=\{\mathcal{E}_{a_1},\dots,\mathcal{E}_{a_M}\}$, where
\begin{equation}
\small
\mathcal{E}_{a_m}=\bigcup_{(i,j)} \mathcal{E}_{a_m}^{(i,j)}.
\end{equation}

\paragraph{Deliberative Intensity Agent (DIA):}
A Deliberative Intensity Agent (DIA) serves as the core reasoning unit for assigning graded contradiction intensity scores. Given an aspect-aligned evidence pair $(e_1^{(j)}, e_2^{(j)}) \in \mathcal{E}_{a_j}$,
the agent functions as a probabilistic mapping that predicts a discrete intensity label $\alpha_j \in \{0,1,2,3\}$ (following the rubric of contradiction intensity \footnote{ Here, label 0 denotes “no valid contradiction” (i.e., the candidate pair does not constitute a genuine contradiction and is filtered out by the validity gate). We will discuss this in detail in Section \ref{Para: validity_gate}.}) and generates a supporting explanation/reason for the assigned label $\rho_j$:
\begin{equation}
(\alpha_j, \rho_j)=g_{\textsc{DIA}}(e_1^{(j)}, e_2^{(j)}, r_i, r_j),
\end{equation}
where $r_i$ and $r_j$ denote the full review contexts. Conditioning on the full context enables the agent to interpret localized evidence spans within the broader evaluative discourse of each reviewer, distinguishing genuine conflict from rhetorical differences. IMPACT employs two DIAs (DIA-A and DIA-B) which share a functional specification but may be instantiated using diverse underlying LLMs to encourage reasoning variance.

\paragraph{Intensity Agreement Checker:}
The Intensity Agreement Checker functions as a deterministic control gate. It compares the agents' initial independent predictions, $\alpha_j^{A}$ and $\alpha_j^{B}$, to determine whether they agree (i.e., $\alpha_j^{A} = \alpha_j^{B}$). If agreement holds, the shared intensity label is accepted directly and propagated to downstream components without further interaction. Conversely, in the event of disagreement, the deliberation protocol is triggered and managed by the Disagreement Orchestrator.

\paragraph{Disagreement Orchestrator:}
The Disagreement Orchestrator (DO) manages structured interaction between DIA-A and DIA-B during disagreement.
Given fixed predictions $\alpha_j^{A} \neq \alpha_j^{B}$, the orchestrator enforces a
\emph{score-locking} constraint:
\begin{equation}
\small
\alpha_{j,t}^{k}=\alpha_{j,0}^{k}, \quad k\in\{A,B\}, \; t\in\{1,\dots,T\}.
\end{equation}

This constraint prevents agents from exhibiting \emph{conformity bias} or drifting toward premature "lazy consensus" \cite{DBLP:conf/icml/Du00TM24, DBLP:conf/emnlp/Liang0JW00Y0T24}. Instead, it compels each agent to deepen its reasoning and generate maximal conflicting evidence, ensuring the Adjudication Agent receives a comprehensive argumentative trace that exposes the nuances of the disagreement. Deliberation proceeds in alternating turns, where agents receive opponent rationales as feedback and respond by grounding arguments in evidence, clarifying intensity criteria, and addressing counter-arguments. The orchestrator maintains the debate history and terminates interaction after a fixed number of rounds.

\paragraph{Adjudication Agent:}
Inspired by LLM as judge \cite{liu-etal-2023-g,DBLP:conf/coling/0010CZGC24}, the Adjudication Agent resolves disagreements by arbitrating over deliberative reasoning rather than re-estimating intensity.
Given the deliberation trace, it selects
\begin{equation}
\small
\alpha_j^{*} \in \{\alpha_j^{A}, \alpha_j^{B}\}
\end{equation}
and produces a consolidated intensity reasoning.

\paragraph{Contradiction Validity Gate (CVG):} \label{Para: validity_gate}
While contradiction intensity is defined over $\{1,2,3\}$, not all extracted aspect-aligned evidence pairs correspond to genuine contradictions. We, therefore, extend the label space with a null label $\alpha = 0$ to jointly model contradiction validity and severity. The Contradiction Validity Gate filters adjudicated outputs based on the final intensity score $\alpha_j^{*}$.
Evidence pairs with $\alpha_j^{*} = 0$ are discarded, while those with $\alpha_j^{*} \ge 1$ are retained. Accordingly, the aspect-specific evidence pool is updated as
\begin{equation}
\small
\mathcal{E}_{a_m}^{*} =
\{(e_1^{(j)}, e_2^{(j)}) \in \mathcal{E}_{a_m} \mid \alpha_j^{*} \ge 1\},
\end{equation}
ensuring that $\mathcal{E}$ contains only adjudicated, valid contradiction evidence with non-zero intensity.

\subsection{TIDE: Teacher--Student Distillation for Evidence-Grounded Intensity Reasoning}

While \textsc{IMPACT} produces high-quality, evidence-grounded contradiction judgments through deliberation and adjudication, its iterative multi-agent process is computationally expensive, limiting large-scale or real-time deployment. To address this limitation, we adopt a teacher--student paradigm, using \textsc{IMPACT} as a \emph{teacher} to generate a high-fidelity synthetic dataset for training a computationally efficient Small Language Model (SLM) \cite{taori2023alpaca,mitra2024orca}.

\paragraph{Automated Data Generation:}
Given a pair of peer reviews $(r_i, r_j)$ of the same paper, the teacher framework identifies a set of contradiction instances $\{c_j\}_{j=1}^{m}$. Each instance $c_j = (e_j, \alpha_j^{*}, \rho_j)$ consists of an aspect-aligned contradiction evidence pair $e_j = (e_{1}^{(j)}, e_{2}^{(j)})$, where $e_{1}^{(j)}$ and $e_{2}^{(j)}$ are verbatim sentences extracted from $r_i$ and $r_j$, respectively, an adjudicated contradiction intensity label $\alpha_j^{*} \in \{1,2,3\}$, and a consolidated intensity reasoning $\rho_j$ produced by IMPACT. Each training example therefore takes the full review pair $(r_i, r_j)$ as input and the corresponding set of contradiction instances as structured supervision. We collect approximately 2{,}000 review pairs from ICLR 2021--2023 via OpenReview\footnote{\url{https://openreview.net/}}. Since IMPACT-P produces higher-quality outputs than large language models such as GPT-5.2~\citep{openai2024gpt4}, Gemini-3 Flash~\citep{anil2023gemini}, LLaMA-4 Maverick~\citep{meta2024llama4}, and existing multi-agent frameworks, we use it to generate contradiction annotations for these review pairs. Statistics of the resulting dataset are shown in Figures~\ref{fig:aspect_dist} and~\ref{fig:intensity_dist}. We split the dataset into 80\% and 20\% for training and validation, respectively, and use the expert-annotated RevCI dataset for testing.




\paragraph{Student Model Training.}
The student SLM is trained to directly model the conditional distribution $p_{\theta}(\{c_j\} \mid r_i, r_j)$ using supervised fine-tuning (SFT) with a next-token prediction objective. To enable parameter-efficient adaptation, we inject Low-Rank Adaptation (LoRA) layers \cite{hu2022lora} into the attention and feed-forward modules of the student model. At inference time, \textsc{TIDE} produces contradiction evidence and the associated intensity score and reasoning in a single forward pass.





\section{Experiments}

\subsection{Implementation Details}

In the multi-agent framework, to eliminate stochastic variation, we disable nucleus and top-$k$ sampling ($p=1.0$, $k=0$) and set the temperature to 0 for all agents. All experiments are run with a fixed random seed = 42, ensuring determinism across repeated runs. Duplicate contradictions are removed using ROUGE-L similarity with a threshold of 0.9.

We evaluate two variants of our framework: \textsc{IMPACT} (Proprietary), abbreviated as \textsc{IMPACT-P}, and \textsc{IMPACT} (Open-Access), abbreviated as \textsc{IMPACT-OA}. Both variants share the same pipeline structure, agent roles, and aspect-conditioned evidence mechanism, and differ only in the choice of underlying models. Details of the specific models used for each agent, along with additional experimental and training settings, are provided in Appendix~\ref{app:impl}.

For our proposed model TIDE, we fine-tune an LLM (Meta-Llama-3-8B-Instruct\footnote{\url{https://huggingface.co/meta-llama/Meta-Llama-3-8B-Instruct}}) using supervised fine-tuning with LoRA adapters. Training is performed for 5 epochs using the AdamW optimizer~\cite{loshchilov2019decoupled} with a fixed learning rate of $5 \times 10^{-5}$, cosine learning rate scheduling, and a warm-up ratio of 0.03. During supervised fine-tuning, only LoRA adapter parameters are updated, while all base model parameters are frozen. LoRA adapters are applied to the attention projection layers (\texttt{q\_proj}, \texttt{k\_proj}, \texttt{v\_proj}, \texttt{o\_proj}) and the feed-forward network projection layers (\texttt{gate\_proj}, \texttt{up\_proj}, \texttt{down\_proj}).

\subsection{Evaluation Metrics}
Contradiction evidence extraction yields unordered sets of instances with variable cardinality and variable-length evidence, making count-based evaluation inadequate. We therefore evaluate evidence overlap using ROUGE-L and enforce a one-to-one alignment between the ground-truth set $G$ and predicted set $P$ via maximum-weight matching (Hungarian algorithm). When $|G| \neq |P|$, we discard weak alignments with ROUGE-L below $\lambda_{\text{match}}$ and treat discarded/unmatched instances as FP/FN. We further report agreement metrics for evidence intensity over the matched pairs. Full details are provided in Appendix~\ref{Appendix:metrics}.


\section{Result and Analysis}
\begin{table*}[!ht]
\centering
\small
\begin{tabular}{llccccc}
\toprule
\textbf{Category} & \textbf{Method} 
& $\mathrm{FNR}\,\downarrow$ 
& $\mathrm{FPR}\,\downarrow$ 
& $\kappa\,\uparrow$ 
& $\rho\,\uparrow$ 
& $\tau\,\uparrow$ \\
\midrule
\multirow{4}{*}{\textbf{Single-agent (CoT)}} 
& GPT-5.2 \cite{openai2024gpt4}
& 0.2935 & 0.3012 & 0.2612 & 0.3679 & 0.3043 \\
& LLaMA-4 Maverick \cite{meta2024llama4}
& 0.2639 & 0.4680 & 0.2348 & 0.3315 & 0.2652 \\
& Gemini-3 Flash \cite{anil2023gemini}
& 0.3741 & 0.2369 & 0.3041 & 0.4523 & 0.3799 \\
& LLaMA-3-8B-Instruct \cite{grattafiori2024llama3}
& 0.3570 & 0.3948 & 0.2488 & 0.3541 & 0.3364 \\
& Qwen-2.5-7B-Instruct \cite{yang2025qwen3} 
& 0.4127 & 0.4534 & 0.1859 & 0.2854 & 0.2283 \\
& ContraSciView \cite{kumar-etal-2023-reviewers} & 0.3920 & 0.3360 & -- & -- & -- \\
\midrule
\multirow{5}{*}{\textbf{Generic multi-agent}} 
& Self-Refine \cite{madaan2023self}      & 0.2750 & 0.2950 & 0.2680 & 0.3810 & 0.3180 \\
& Debate \cite{DBLP:conf/acl/KimKY24}             & 0.2870 & 0.2820 & 0.2550 & 0.3650 & 0.3010 \\
& MAD \cite{liang-etal-2024-encouraging}            & 0.2690 & 0.2760 & 0.2720 & 0.3890 & 0.3270 \\
& ChatEval \cite{chan2023chateval}         & 0.2580 & 0.2640 & 0.2810 & 0.4020 & 0.3410 \\
& CourtEval \cite{kumar-etal-2025-courteval}        & 0.2520 & 0.2590 & 0.2860 & 0.4100 & 0.3490 \\
\midrule
\multirow{3}{*}{\textbf{Our Proposed}} 
& \textbf{IMPACT-OA} & 0.2390 & 0.2287 & 0.3270 & 0.4783 & 0.4421 \\
& \textbf{IMPACT-P}  & \textbf{0.1901} & \textbf{0.1613} & \textbf{0.3862} & \textbf{0.6193} & \textbf{0.5826} \\
& \textbf{TIDE}      & 0.3771 & 0.3048 & 0.2202 & 0.3793 & 0.3549 \\
\bottomrule
\end{tabular}

\caption{Overall comparison of contradiction detection and intensity agreement.
We compute $\mathrm{FNR}$ and $\mathrm{FPR}$ at the \emph{review-pair level} over full test set (positive = contradiction present; negative = no contradiction), where $\mathrm{FPR}=\frac{\mathrm{FP}}{\mathrm{FP}+\mathrm{TN}}$ and $\mathrm{FNR}=\frac{\mathrm{FN}}{\mathrm{TP}+\mathrm{FN}}$.
We compute $\kappa$, $\rho$, and $\tau$ over matched contradiction evidence pairs (i.e., where intensity is defined). Here, CoT means using chain of thought prompting.}
\label{tab:baseline_comparison}
\end{table*}

\subsection{Main Result} We report the results of our proposed multi-agent framework, \textsc{IMPACT}, under two settings: \textsc{IMPACT-P} and \textsc{IMPACT-OA}, along with our proposed SLM, \textsc{TIDE}, in Table~\ref{tab:baseline_comparison}. We find that the \textsc{IMPACT} framework advances the state of the art on both contradiction evidence detection and intensity estimation. In particular, \textsc{IMPACT-P} and \textsc{IMPACT-OA} achieve the lowest error rates and highest agreement among all baselines, including single-agent models, generic multi-agent frameworks, and the prior task-specific system ContraSciView. Compared to the strongest baseline, \textbf{CourtEval}, \textsc{IMPACT-P} reduces the average detection error (avg.\ FNR and FPR) by \textbf{31.2\%} and improves the average agreement score (avg.\ $\kappa$, $\rho$, and $\tau$) by \textbf{52.0\%}. Even without proprietary models, \textsc{IMPACT-OA} achieves an \textbf{8.5\%} reduction in average detection error and a \textbf{19.4\%} improvement in average agreement over CourtEval, indicating that the gains primarily stem from the framework design rather than model scale. Finally, we report results for \textsc{TIDE}. As shown in Table~\ref{tab:baseline_comparison}, \textsc{TIDE} attains lower average evidence detection error than LLaMA-4 Maverick, with a \textbf{6.8\%} reduction in avg.\ FNR/FPR, and achieves higher average agreement with human annotations than GPT-5.2, improving avg.\ $(\kappa,\rho,\tau)$ by \textbf{2.3\%}. For fair comparison, all baseline LLMs are evaluated using the same prompting template provided in Appendix~\ref{app:llm_prompts}. However, \textsc{TIDE} does not uniformly outperform all baselines across individual metrics; instead, it provides a favorable trade-off between false positive rate and human-alignment metrics. Additionally, \textsc{TIDE} requires only a single forward pass, making it substantially more efficient than multi-agent frameworks such as \textsc{IMPACT}. These results indicate that the structured supervision provided by \textsc{IMPACT} transfers fine-grained intensity reasoning to smaller models. We additionally perform a paired bootstrap test over the RevCI test set and find that the improvement in composite agreement score is statistically significant ($p < 0.05$).


\subsection{Effect of Discussion Rounds}
\begin{figure}[ht]
  \fbox{%
    \includegraphics[width=\columnwidth]{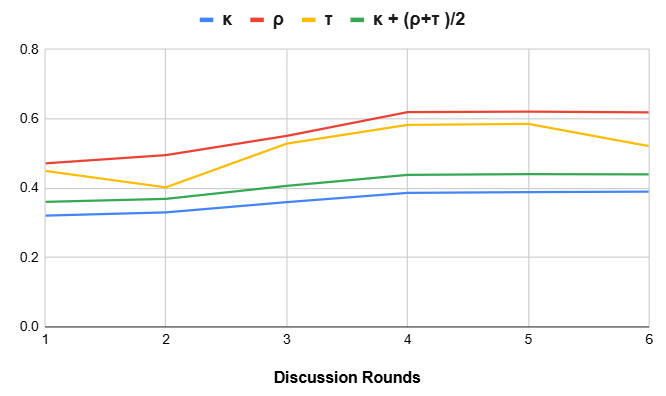}
  }
  \caption{The figure shows the trends of the intensity agreement metrics $\kappa$, $\rho$, $\tau$ and their composite score $C=\kappa+(\rho+\tau)/2$ over successive discussion rounds.}
  \label{fig:experiments}
\end{figure}

To quantify the impact of deliberation depth, we vary the number of discussion rounds $D$ in our multi-agent deliberation framework, and let a downstream judge produce the final intensity label conditioned on the resulting dialogue. Figure~\ref{fig:experiments} reports agreement with human annotations using Cohen’s $\kappa$, which measures categorical agreement corrected for chance; Spearman’s $\rho$, which captures rank-order correlation; Kendall’s $\tau$, which measures pairwise ranking consistency. These metrics capture complementary aspects of agreement: $\kappa$ evaluates exact label matching, while $\rho$ and $\tau$ assess ordinal consistency between predictions and human judgments. To summarize overall performance, we define a composite score $C=\kappa+(\rho+\tau)/2$ , which balances categorical agreement with ordinal consistency.

Overall, increasing $D$ yields consistent improvements up to a small number of rounds, after which gains saturate. The composite score improves from $C=0.3608$ with a single round to $0.3691$ with two rounds (+2.3\%). Increasing to three rounds yields a pronounced jump to $0.4068$ (+9.26\%), the largest relative gain, accompanied by concurrent improvements across all constituent metrics. Moving from three to four rounds provides an additional but attenuated improvement (+7.76\%). Beyond four rounds, the marginal benefit becomes negligible: increasing from four to five rounds yields only a +0.57\% change, while extending to six rounds slightly decreases performance (-0.27\%). Taken together, these results suggest that most of the benefit of deliberation is achieved within the first four rounds, indicating convergence and motivating $D=4$ as a practical operating point. We also discuss false positive and false negative rates in Appendix~\ref{Appendix:fnr}.

\subsection{Ablation Study of IMPACT} \label{sec:ablation_aspect_conditioning}

\begin{table*}[t]
\centering
\small
\begin{tabular}{cccccccccc}
\toprule
\textbf{DO} & \textbf{ACEA} & \textbf{IEx} & \textbf{IS} & \textbf{CVG}
& \textbf{FNR}$\downarrow$ & \textbf{FPR}$\downarrow$ 
& \textbf{$\kappa$}$\uparrow$ & \textbf{$\rho$}$\uparrow$ & \textbf{$\tau$}$\uparrow$ \\
\midrule

$\times$ & $\times$ & $\times$ & $\times$ & $\times$
& 0.2969 & 0.3661 & -- & -- & -- \\

$\times$ & \checkmark & $\times$ & $\times$ & $\times$
& 0.1092 & 0.5120 & -- & -- & -- \\

$\times$ & $\times$ & $\times$ & \checkmark & $\times$
& 0.3570 & 0.3948 & 0.2488 & 0.3541 & 0.3364 \\

$\times$ & $\times$ & \checkmark & \checkmark & $\times$
& 0.3293 & 0.3346 & 0.3392 & 0.5134 & 0.4219 \\

$\times$ & \checkmark & \checkmark & \checkmark & \checkmark
& 0.1953 & 0.2614 & 0.3115 & 0.4574 & 0.4308 \\

\checkmark & \checkmark & \checkmark & \checkmark & \checkmark
& \textbf{0.1901} & \textbf{0.1613} & \textbf{0.3862} & \textbf{0.6193} & \textbf{0.5826} \\

\bottomrule
\end{tabular}
\caption{
Ablation of \textsc{IMPACT} components.
DO: Disagreement Orchestrator; ACEA: Aspect Conditioning Evidence Agent; IEx: intensity examples; IS: intensity scoring; CVG: Contradiction Validity Gate.
}
\label{tab:ablation_short}
\end{table*}

Table~\ref{tab:ablation_short} reports that without aspect conditioning, the model exhibits a relatively high FNR of 0.2969, indicating that a substantial fraction of true contradictions are missed. Introducing aspect conditioning reduces the FNR to 0.1092, corresponding to a 63\% relative reduction and an increase in recall from 0.703 to 0.891. This result indicates that explicit aspect conditioning substantially improves the model’s ability to identify true contradictions.

At the same time, aspect conditioning (AC) increases the FPR from 0.3661 to 0.5120. This trade-off is expected: conditioning the model on a specific aspect reduces ambiguity about relevance, but can also lead to over-identification, which can be filtered out later using other methods. These results indicate that aspect conditioning helps reduce the FNR.

Table~\ref{tab:ablation_short} also shows that when we use a single agent\footnote{We prompt a single LLM for generating contradictions and assigning them an intensity score (1-3)}. For intensity scoring with examples (IS+IEx), the FNR and FPR show an average reduction of 11.7\% relative to scoring without intensity examples, along with a 35.68\% increase in average agreement metrics. These results demonstrate the importance of intensity examples in the framework understanding what each intensity score means.

We then combine AC and IS+IEx to achieve a low FNR and in order to filter out the increased false positives incorporated by aspect conditioning we use CVG. Thus by combining AC, IS+IEx and CVG we reduce the FNR and FPR by 40.69\% and 21.87\% respectively, compared to the IS+IEx agent. This shows that this component is important part of our framework where AC increases total contradictions detected reducing FNR and CVG filtering out false positives reducing FPR.

We add DO, DIAs, and an Adjudication agent, which results in a further reduction of the false positive rate (FPR) by 38.29\% and an increase in the average agreement score by 32.37\%. These results indicate that iterative reasoning and discussion help further remove incorrect contradiction evidence and improve the agreement of intensity assignments.

\subsection{Ablation Study of TIDE}

\begin{table*}[t]
\centering
\small
\begin{tabular}{cccccccc}
\hline
\textbf{FT} & \textbf{IS} & \textbf{IR} & \textbf{FNR} & \textbf{FPR} & \textbf{$\kappa$} & \textbf{$\rho$} & \textbf{$\tau$} \\
\hline
$\times$ & $\times$ & $\times$ & 0.6641 & 0.4699 & -- & -- & -- \\
\checkmark & $\times$ & $\times$ & 0.4162 & 0.3390 & -- & -- & -- \\
$\times$ & \checkmark & $\times$ & 0.5279 & 0.4529 & -0.0250 & 0.1384 & 0.1231 \\
\checkmark & \checkmark & $\times$ & 0.4111 & 0.3555 & 0.1699 & 0.2980 & 0.2530 \\
\checkmark & \checkmark & \checkmark & \textbf{0.3771} & \textbf{0.3048} & \textbf{0.2202} & \textbf{0.3793} & \textbf{0.3549} \\
\hline
\end{tabular}
\caption{Ablation of TIDE components. FT: Finetuned; IS: Intensity Scoring; IR:Intensity Reasoning.}
\label{tab: ablation_slm}
\end{table*}

Table~\ref{tab: ablation_slm} reports the ablation study of different components of TIDE. When prompted with identical instructions to generate contradictions from paired reviews in the test set, the LLaMA-3.1-8B-Instruct base model ($\times$FT (Finetuned), $\times$IS (Intensity Scorer), $\times$IR (Intensity Reasoning)) exhibits high error rates, with an FNR of 0.6641 and an FPR of 0.4699, as shown in Table~\ref{tab: ablation_slm}. Fine-tuning the model without incorporating intensity supervision ($\checkmark$FT, $\times$IS, $\times$IR) substantially reduces these errors, achieving an average relative reduction of 33.40\% in FPR and FNR. 

Introducing explicit intensity scoring supervision during fine-tuning ($\checkmark$FT, $\checkmark$IS, $\times$IR) further improves performance, lowering the FNR to 0.4111 and the FPR to 0.3555, while also enabling meaningful alignment with human intensity annotations, reflected in a 67.2\% increase in average agreement metrics compared to the base model with intensity scoring. These results indicate that incorporating intensity scoring supervision helps the LLM better distinguish between contradictions and non-contradictions.

Finally, incorporating explicit intensity reasoning during training ($\checkmark$FT, $\checkmark$IS, $\checkmark$IR) yields the best overall result. This full TIDE configuration achieves the lowest error rates (FNR 0.3771, FPR 0.3048) and the strongest alignment with human judgments ($\kappa = 0.2202$, $\rho = 0.3793$, $\tau = 0.3549$), demonstrating an average reduction in FPR and FNR values by 11.04\% as compared to the finetuned model without intensity reasoning. Thus, training on intensity reasoning helps LLMs understand the logic behind the intensity scores and better distinguish between different scores, increasing the average score of alignment metrics.

\subsection{Human Evaluation}
Our model mainly fails due to (i) misreading vague/scalar evaluative language as contradictions and (ii) confusing which aspect is being discussed when sentences mention multiple aspects. We discuss the error analysis and the case study in detail in Appendices \ref{Appendix_error} and \ref{Appendix: case_study}, respectively.


\section{Conclusion and Future Work}
We study reviewer disagreement through fine-grained contradiction analysis over full scientific peer reviews, introducing a new task and the \textbf{RevCI} benchmark for evidence-grounded contradiction identification and graded intensity modeling. We propose \textbf{IMPACT}, a structured multi-agent framework for contradiction detection and intensity estimation, and \textbf{TIDE}, a distilled small language model that achieves competitive performance with substantially lower inference cost. Our results show that reviewer disagreement is often fine-grained and context-dependent, making binary contradiction detection over isolated sentence pairs insufficient, and that IMPACT consistently outperforms strong single-agent and generic multi-agent baselines, including settings without proprietary models, with gains driven by task-specific deliberative design rather than model scale. Finally, TIDE demonstrates that this fine-grained intensity reasoning can be effectively distilled into an efficient model while maintaining strong alignment with human annotations, and future work will explore generalization beyond the computer science domain and extension to multi-reviewer disagreement.

\section*{Limitation}
We restrict our experiments to the most frequent aspect categories in peer reviews, i.e., \textit{Motivation, Clarity, Soundness, Substance, Originality}, and \textit{Meaningful Comparison}. These aspects account for the majority of explicit contradiction instances in our annotated data, enabling reliable annotation and stable evaluation. While prior work identifies additional aspect categories \cite{lu-etal-2025-identifying}, many occur too sparsely to support robust modeling. Our framework supports extensibility: in \textsc{IMPACT}, new aspects can be incorporated by updating the ACEA prompt with the corresponding aspect name and definition, without retraining. In contrast, extending \textsc{TIDE} to unseen aspect categories requires retraining, as the student model learns aspect-specific decision boundaries during distillation; however, this retraining is lightweight due to parameter-efficient fine-tuning. Our dataset is constructed from peer reviews primarily from ICLR and NeurIPS, and may not fully capture the diversity of review practices across venues, domains, or reviewer expertise levels. Additionally, the LLM-based pre-filtering used to select candidate review pairs may bias the dataset toward more explicit contradictions, potentially under-representing subtle or implicit disagreements. However, all final annotations are produced by expert annotators using full review context, ensuring that the benchmark reflects human judgment rather than model-generated labels. We leave broader cross-venue validation and more diverse sampling strategies to future work.

\section*{Ethics}
This work uses peer-review texts derived from publicly available sources and synthetic data generated by our models; all data are anonymized where applicable and contain no personal, sensitive, or identifying information. The proposed system is intended solely as an assistive tool for editors and area chairs to help identify potential contradictions in lengthy and complex reviews, and it is not designed to automate editorial decisions or to evaluate authors or reviewers. As with any AI-based system, the model is not perfectly accurate and may produce false positives or false negatives; therefore, all outputs must be interpreted with human oversight and verified using standard editorial judgment. For example, when one reviewer states, ``The paper does not have any new findings,'' while another notes, ``The paper is somewhat novel,'' such disagreement may not be immediately apparent in long reviews, and the system aims only to surface such cases to prompt further inspection. Failure to identify a contradiction does not imply that none exists, and contradictions identified outside the system’s outputs should still be handled according to established review guidelines. We caution against misuse of the system for monitoring or penalizing reviewers or for exposing model outputs directly to authors, and emphasize that it should be deployed only for internal editorial support with appropriate safeguards. We used an AI-based writing assistant (ChatGPT) for minor grammatical corrections and proofreading.

\section*{Acknowledgement}
Sandeep Kumar acknowledges the Prime Minister Research Fellowship (PMRF) program of the Govt of India for its support.

\bibliography{custom}

@article{DBLP:journals/arist/Bornmann11,
  author       = {Lutz Bornmann},
  title        = {Scientific peer review},
  journal      = {Annu. Rev. Inf. Sci. Technol.},
  volume       = {45},
  number       = {1},
  pages        = {197--245},
  year         = {2011},
  url          = {https://doi.org/10.1002/aris.2011.1440450112},
  doi          = {10.1002/ARIS.2011.1440450112},
  bibsource    = {dblp computer science bibliography, https://dblp.org}
}

@inproceedings{rogers-augenstein-2020-improve,
    title = "What Can We Do to Improve Peer Review in {NLP}?",
    author = "Rogers, Anna  and
      Augenstein, Isabelle",
    editor = "Cohn, Trevor  and
      He, Yulan  and
      Liu, Yang",
    booktitle = "Findings of the Association for Computational Linguistics: EMNLP 2020",
    month = nov,
    year = "2020",
    address = "Online",
    publisher = "Association for Computational Linguistics",
    url = "https://aclanthology.org/2020.findings-emnlp.112/",
    doi = "10.18653/v1/2020.findings-emnlp.112",
    pages = "1256--1262"
}

@inproceedings{10.1145/3209978.3210056,
author = {Wang, Ke and Wan, Xiaojun},
title = {Sentiment Analysis of Peer Review Texts for Scholarly Papers},
year = {2018},
isbn = {9781450356572},
publisher = {Association for Computing Machinery},
address = {New York, NY, USA},
url = {https://doi.org/10.1145/3209978.3210056},
doi = {10.1145/3209978.3210056},
booktitle = {The 41st International ACM SIGIR Conference on Research \& Development in Information Retrieval},
pages = {175–184},
numpages = {10},
location = {Ann Arbor, MI, USA},
series = {SIGIR '18}
}

@article{980f5632f0664f9681dea65f7a810bff,
title = "Peer review quality and transparency of the peer-review process in open access and subscription journals",
author = "J.M. Wicherts",
year = "2016",
doi = "10.1371/journal.pone.0147913",
language = "English",
volume = "11",
journal = "PLOS ONE",
issn = "1932-6203",
publisher = "PUBLIC LIBRARY SCIENCE",
number = "1",
}

@article{fbdf6c99376b42789f9a8e5305d8afed,
title = "Empowering peer reviewers with a checklist to improve transparency",
author = "Timothy Parker and Simon Griffith and Judith Bronstein and Fiona Fidler and Susan Foster and Hannah Fraser and Wolfgang Forstmeier and Jessica Gurevitch and Julia Koricheva and Ralf Seppelt and Morgan Tingley and Shinichi Nakagawa",
year = "2018",
month = may,
day = "22",
doi = "10.1038/s41559-018-0545-z",
language = "English",
volume = "2",
pages = "929--935",
journal = "Nature Ecology \& Evolution",
issn = "2397-334X",
publisher = "Nature Publishing Group",
}

@article{DBLP:journals/pacmhci/StelmakhSSD21,
  author       = {Ivan Stelmakh and
                  Nihar B. Shah and
                  Aarti Singh and
                  Hal Daum{\'{e}} III},
  title        = {Prior and Prejudice: The Novice Reviewers' Bias against Resubmissions
                  in Conference Peer Review},
  journal      = {Proc. {ACM} Hum. Comput. Interact.},
  volume       = {5},
  number       = {{CSCW1}},
  pages        = {75:1--75:17},
  year         = {2021},
  url          = {https://doi.org/10.1145/3449149},
  doi          = {10.1145/3449149},
  bibsource    = {dblp computer science bibliography, https://dblp.org}
}

@article{DBLP:journals/cacm/FreyneCSC10,
  author       = {Jill Freyne and
                  Lorcan Coyle and
                  Barry Smyth and
                  Padraig Cunningham},
  title        = {Relative status of journal and conference publications in computer
                  science},
  journal      = {Commun. {ACM}},
  volume       = {53},
  number       = {11},
  pages        = {124--132},
  year         = {2010},
  url          = {https://doi.org/10.1145/1839676.1839701},
  doi          = {10.1145/1839676.1839701},
  bibsource    = {dblp computer science bibliography, https://dblp.org}
}

@article{DBLP:journals/scientometrics/BrezisB20,
  author       = {Elise S. Brezis and
                  Aliaksandr Birukou},
  title        = {Arbitrariness in the peer review process},
  journal      = {Scientometrics},
  volume       = {123},
  number       = {1},
  pages        = {393--411},
  year         = {2020},
  url          = {https://doi.org/10.1007/s11192-020-03348-1},
  doi          = {10.1007/s11192-020-03348-1},
  bibsource    = {dblp computer science bibliography, https://dblp.org}
}

@article{DBLP:journals/cacm/LangfordG15,
  author       = {John Langford and
                  Mark Guzdial},
  title        = {The arbitrariness of reviews, and advice for school administrators},
  journal      = {Commun. {ACM}},
  volume       = {58},
  number       = {4},
  pages        = {12--13},
  year         = {2015},
  url          = {https://doi.org/10.1145/2732417},
  doi          = {10.1145/2732417},
  bibsource    = {dblp computer science bibliography, https://dblp.org}
}

@article{peer_review_stress,
author = {Kelly, Jacalyn and Sadeghieh, Tara and Adeli, Khosrow},
year = {2014},
month = {10},
pages = {227-43},
title = {Peer Review in Scientific Publications: Benefits, Critiques, \& A Survival Guide},
volume = {25},
journal = {EJIFCC}
}

@article{10.1093/biosci/bix034,
    author = {Gropp, Robert E. and Glisson, Scott and Gallo, Stephen and Thompson, Lisa},
    title = "{Peer Review: A System under Stress}",
    journal = {BioScience},
    volume = {67},
    number = {5},
    pages = {407-410},
    year = {2017},
    month = {05},
    issn = {0006-3568},
    doi = {10.1093/biosci/bix034},
    url = {https://doi.org/10.1093/biosci/bix034},
    eprint = {https://academic.oup.com/bioscience/article-pdf/67/5/407/14172931/bix034.pdf},
}

@proceedings{emnlp2025chairs,
    title = "Proceedings of the 2025 Conference on Empirical Methods in Natural Language Processing",
    editor = "Christodoulopoulos, Christos  and
      Chakraborty, Tanmoy  and
      Rose, Carolyn  and
      Peng, Violet",
    month = nov,
    year = "2025",
    address = "Suzhou, China",
    publisher = "Association for Computational Linguistics",
    url = "https://aclanthology.org/2025.emnlp-main.0/",
    doi = "10.18653/v1/2025.emnlp-main.0",
    ISBN = "979-8-89176-332-6"
}

@inproceedings{liang2024monitoring,
author = {Liang, Weixin and Izzo, Zachary and Zhang, Yaohui and Lepp, Haley and Cao, Hancheng and Zhao, Xuandong and Chen, Lingjiao and Ye, Haotian and Liu, Sheng and Huang, Zhi and McFarland, Daniel A. and Zou, James Y.},
title = {Monitoring AI-modified content at scale: a case study on the impact of ChatGPT on AI conference peer reviews},
year = {2024},
publisher = {JMLR.org},
booktitle = {Proceedings of the 41st International Conference on Machine Learning},
articleno = {1192},
numpages = {46},
location = {Vienna, Austria},
series = {ICML'24}
}

@article{Mishra_2025, title={Challenges in the Peer-review process}, volume={5}, url={https://penerbitadm.pubmedia.id/index.php/JIM/article/view/2570}, DOI={10.53697/jim.v5i1.2570}, abstractNote={&amp;lt;p&amp;gt;The peer-review process, considered as the backbone of academic publishing, faces many challenges that undermine its reliability and effectiveness. These issues affect the accuracy of published research and contribute to frustration among authors and reviewers. This study delves into these challenges through qualitative interviews with 10 academic stakeholders, including researchers, reviewers, editors, and the editor-in-chief. The primary focus of the research is to uncover the key issues impacting the peer-review system and to propose practical solutions for addressing them. Using thematic analysis, the study identifies several persistent issues, including the overwhelming workload faced by reviewers, delays in providing feedback, and the influence of personal biases on review outcomes. These factors lead to inconsistent and sometimes unreliable evaluations of research, which can hinder the publication process. Moreover, the lack of standardised review criteria further exacerbates the situation, with different reviewers applying varying standards to the same manuscript. Such inconsistencies compromise the quality and speed of the review process, resulting in significant challenges for both authors and reviewers. The paper proposes several solutions to improve the peer-review system in light of these findings. By addressing these issues, the study contributes to ongoing efforts to enhance the effectiveness of the peer-review system and ensure its continued relevance in the rapidly evolving landscape of academic publishing.&amp;lt;/p&amp;gt;}, number={1}, journal={Journal of Indonesian Management}, author={Mishra, Udgam}, year={2025}, month={Mar.}, pages={17} }

@article{taori2023alpaca,
  title={Alpaca: A strong, replicable instruction-following model},
  author={Taori, Rohan and Gulrajani, Ishaan and Zhang, Tianyi and Dubois, Yann and Li, Xuechen and Guestrin, Carlos and Liang, Percy and Hashimoto, Tatsunori B},
  journal={Stanford Center for Research on Foundation Models. https://crfm. stanford. edu/2023/03/13/alpaca. html},
  volume={3},
  number={6},
  pages={7},
  year={2023}
}

@article{mitra2024orca,
  title={Orca-math: Unlocking the potential of slms in grade school math},
  author={Mitra, Arindam and Khanpour, Hamed and Rosset, Corby and Awadallah, Ahmed},
  journal={arXiv preprint arXiv:2402.14830},
  year={2024}
}

@article{hu2022lora,
  title={Lora: Low-rank adaptation of large language models.},
  author={Hu, Edward J and Shen, Yelong and Wallis, Phillip and Allen-Zhu, Zeyuan and Li, Yuanzhi and Wang, Shean and Wang, Lu and Chen, Weizhu and others},
  journal={ICLR},
  volume={1},
  number={2},
  pages={3},
  year={2022}
}

@article{kendall1938new,
  title={A new measure of rank correlation},
  author={Kendall, Maurice G},
  journal={Biometrika},
  volume={30},
  number={1-2},
  pages={81--93},
  year={1938},
  publisher={Oxford University Press}
}

@article{spearman1961proof,
    author = {Spearman, C},
    title = {The proof and measurement of association between two things},
    journal = {International Journal of Epidemiology},
    volume = {39},
    number = {5},
    pages = {1137-1150},
    year = {2010},
    month = {10},
    issn = {0300-5771},
    doi = {10.1093/ije/dyq191},
    url = {https://doi.org/10.1093/ije/dyq191},
    eprint = {https://academic.oup.com/ije/article-pdf/39/5/1137/18481215/dyq191.pdf},
}

@article{cohen1960coefficient,
  title={A coefficient of agreement for nominal scales},
  author={Cohen, Jacob},
  journal={Educational and psychological measurement},
  volume={20},
  number={1},
  pages={37--46},
  year={1960},
  publisher={Sage Publications Sage CA: Thousand Oaks, CA}
}

@article{DBLP:journals/corr/abs-2102-00176,
  author    = {Weizhe Yuan and
               Pengfei Liu and
               Graham Neubig},
  title     = {Can We Automate Scientific Reviewing?},
  journal   = {CoRR},
  volume    = {abs/2102.00176},
  year      = {2021},
  url       = {https://arxiv.org/abs/2102.00176},
  archivePrefix = {arXiv},
  eprint    = {2102.00176},
  bibsource = {dblp computer science bibliography, https://dblp.org}
}

@inproceedings{loshchilov2019decoupled,
  title     = {Decoupled Weight Decay Regularization},
  author    = {Loshchilov, Ilya and Hutter, Frank},
  booktitle = {International Conference on Learning Representations (ICLR)},
  year      = {2019}
}

@article{scipy,
  title   = {SciPy 1.0: Fundamental Algorithms for Scientific Computing in Python},
  author  = {Virtanen, Pauli and others},
  journal = {Nature Methods},
  volume  = {17},
  pages   = {261--272},
  year    = {2020},
  doi     = {10.1038/s41592-019-0686-2}
}

@inproceedings{DBLP:conf/icml/Du00TM24,
  author       = {Yilun Du and
                  Shuang Li and
                  Antonio Torralba and
                  Joshua B. Tenenbaum and
                  Igor Mordatch},
  title        = {Improving Factuality and Reasoning in Language Models through Multiagent
                  Debate},
  booktitle    = {Forty-first International Conference on Machine Learning, {ICML} 2024,
                  Vienna, Austria, July 21-27, 2024},
  publisher    = {OpenReview.net},
  year         = {2024},
  url          = {https://openreview.net/forum?id=zj7YuTE4t8},
  bibsource    = {dblp computer science bibliography, https://dblp.org}
}

@inproceedings{DBLP:conf/emnlp/Liang0JW00Y0T24,
  author       = {Tian Liang and
                  Zhiwei He and
                  Wenxiang Jiao and
                  Xing Wang and
                  Yan Wang and
                  Rui Wang and
                  Yujiu Yang and
                  Shuming Shi and
                  Zhaopeng Tu},
  editor       = {Yaser Al{-}Onaizan and
                  Mohit Bansal and
                  Yun{-}Nung Chen},
  title        = {Encouraging Divergent Thinking in Large Language Models through Multi-Agent
                  Debate},
  booktitle    = {Proceedings of the 2024 Conference on Empirical Methods in Natural
                  Language Processing, {EMNLP} 2024, Miami, FL, USA, November 12-16,
                  2024},
  pages        = {17889--17904},
  publisher    = {Association for Computational Linguistics},
  year         = {2024},
  url          = {https://doi.org/10.18653/v1/2024.emnlp-main.992},
  doi          = {10.18653/V1/2024.EMNLP-MAIN.992},
  bibsource    = {dblp computer science bibliography, https://dblp.org}
}

@article{madaan2023self,
  title={Self-refine: Iterative refinement with self-feedback},
  author={Madaan, Aman and Tandon, Niket and Gupta, Prakhar and Hallinan, Skyler and Gao, Luyu and Wiegreffe, Sarah and Alon, Uri and Dziri, Nouha and Prabhumoye, Shrimai and Yang, Yiming and others},
  journal={Advances in Neural Information Processing Systems},
  volume={36},
  pages={46534--46594},
  year={2023}
}

@inproceedings{DBLP:conf/acl/KimKY24,
  author       = {Alex Kim and
                  Keonwoo Kim and
                  Sangwon Yoon},
  editor       = {Lun{-}Wei Ku and
                  Andre Martins and
                  Vivek Srikumar},
  title        = {{DEBATE:} Devil's Advocate-Based Assessment and Text Evaluation},
  booktitle    = {Findings of the Association for Computational Linguistics, {ACL} 2024,
                  Bangkok, Thailand and virtual meeting, August 11-16, 2024},
  pages        = {1885--1897},
  publisher    = {Association for Computational Linguistics},
  year         = {2024},
  url          = {https://doi.org/10.18653/v1/2024.findings-acl.112},
  doi          = {10.18653/V1/2024.FINDINGS-ACL.112},
  bibsource    = {dblp computer science bibliography, https://dblp.org}
}

@inproceedings{liang-etal-2024-encouraging,
    title = "Encouraging Divergent Thinking in Large Language Models through Multi-Agent Debate",
    author = "Liang, Tian  and
      He, Zhiwei  and
      Jiao, Wenxiang  and
      Wang, Xing  and
      Wang, Yan  and
      Wang, Rui  and
      Yang, Yujiu  and
      Shi, Shuming  and
      Tu, Zhaopeng",
    editor = "Al-Onaizan, Yaser  and
      Bansal, Mohit  and
      Chen, Yun-Nung",
    booktitle = "Proceedings of the 2024 Conference on Empirical Methods in Natural Language Processing",
    month = nov,
    year = "2024",
    address = "Miami, Florida, USA",
    publisher = "Association for Computational Linguistics",
    url = "https://aclanthology.org/2024.emnlp-main.992/",
    doi = "10.18653/v1/2024.emnlp-main.992",
    pages = "17889--17904"
}

@article{chan2023chateval,
  title={Chateval: Towards better llm-based evaluators through multi-agent debate},
  author={Chan, Chi-Min and Chen, Weize and Su, Yusheng and Yu, Jianxuan and Xue, Wei and Zhang, Shanghang and Fu, Jie and Liu, Zhiyuan},
  journal={arXiv preprint arXiv:2308.07201},
  year={2023}
}

@inproceedings{kumar-etal-2025-courteval,
    title = "{C}ourt{E}val: A Courtroom-Based Multi-Agent Evaluation Framework",
    author = "Kumar, Sandeep  and
      Nargund, Abhijit A  and
      Sridhar, Vivek",
    editor = "Che, Wanxiang  and
      Nabende, Joyce  and
      Shutova, Ekaterina  and
      Pilehvar, Mohammad Taher",
    booktitle = "Findings of the Association for Computational Linguistics: ACL 2025",
    month = jul,
    year = "2025",
    address = "Vienna, Austria",
    publisher = "Association for Computational Linguistics",
    url = "https://aclanthology.org/2025.findings-acl.1327/",
    doi = "10.18653/v1/2025.findings-acl.1327",
    pages = "25875--25887",
    ISBN = "979-8-89176-256-5"
}

@inproceedings{kumar-etal-2023-reviewers,
    title = "When Reviewers Lock Horns: Finding Disagreements in Scientific Peer Reviews",
    author = "Kumar, Sandeep  and
      Ghosal, Tirthankar  and
      Ekbal, Asif",
    editor = "Bouamor, Houda  and
      Pino, Juan  and
      Bali, Kalika",
    booktitle = "Proceedings of the 2023 Conference on Empirical Methods in Natural Language Processing",
    month = dec,
    year = "2023",
    address = "Singapore",
    publisher = "Association for Computational Linguistics",
    url = "https://aclanthology.org/2023.emnlp-main.1038/",
    doi = "10.18653/v1/2023.emnlp-main.1038",
    pages = "16693--16704"
}

@article{fabbri2020summeval,
  title={SummEval: Re-evaluating Summarization Evaluation},
  author={Fabbri, Alexander R and Kry{\'s}ci{\'n}ski, Wojciech and McCann, Bryan and Xiong, Caiming and Socher, Richard and Radev, Dragomir},
  journal={arXiv preprint arXiv:2007.12626},
  year={2020}
}

@inproceedings{liu-etal-2023-g,
    title = "{G}-Eval: {NLG} Evaluation using Gpt-4 with Better Human Alignment",
    author = "Liu, Yang  and
      Iter, Dan  and
      Xu, Yichong  and
      Wang, Shuohang  and
      Xu, Ruochen  and
      Zhu, Chenguang",
    editor = "Bouamor, Houda  and
      Pino, Juan  and
      Bali, Kalika",
    booktitle = "Proceedings of the 2023 Conference on Empirical Methods in Natural Language Processing",
    month = dec,
    year = "2023",
    address = "Singapore",
    publisher = "Association for Computational Linguistics",
    url = "https://aclanthology.org/2023.emnlp-main.153/",
    doi = "10.18653/v1/2023.emnlp-main.153",
    pages = "2511--2522"
}

@inproceedings{DBLP:conf/coling/0010CZGC24,
  author       = {Rui Mao and
                  Guanyi Chen and
                  Xulang Zhang and
                  Frank Guerin and
                  Erik Cambria},
  editor       = {Nicoletta Calzolari and
                  Min{-}Yen Kan and
                  V{\'{e}}ronique Hoste and
                  Alessandro Lenci and
                  Sakriani Sakti and
                  Nianwen Xue},
  title        = {GPTEval: {A} Survey on Assessments of ChatGPT and {GPT-4}},
  booktitle    = {Proceedings of the 2024 Joint International Conference on Computational
                  Linguistics, Language Resources and Evaluation, {LREC/COLING} 2024,
                  20-25 May, 2024, Torino, Italy},
  pages        = {7844--7866},
  publisher    = {{ELRA} and {ICCL}},
  year         = {2024},
  url          = {https://aclanthology.org/2024.lrec-main.693},
  bibsource    = {dblp computer science bibliography, https://dblp.org}
}

@inproceedings{lu-etal-2025-identifying,
    title = "Identifying Aspects in Peer Reviews",
    author = "Lu, Sheng  and
      Kuznetsov, Ilia  and
      Gurevych, Iryna",
    editor = "Christodoulopoulos, Christos  and
      Chakraborty, Tanmoy  and
      Rose, Carolyn  and
      Peng, Violet",
    booktitle = "Findings of the Association for Computational Linguistics: EMNLP 2025",
    month = nov,
    year = "2025",
    address = "Suzhou, China",
    publisher = "Association for Computational Linguistics",
    url = "https://aclanthology.org/2025.findings-emnlp.326/",
    doi = "10.18653/v1/2025.findings-emnlp.326",
    pages = "6145--6167",
    ISBN = "979-8-89176-335-7"
}

@article{bornmann2010usefulness,
  title={The usefulness of peer review for selecting manuscripts for publication: a utility analysis taking as an example a high-impact journal},
  author={Bornmann, Lutz and Daniel, Hans-Dieter},
  journal={PloS one},
  volume={5},
  number={6},
  pages={e11344},
  year={2010},
  publisher={Public Library of Science San Francisco, USA}
}

@inproceedings{kang-etal-2018-dataset,
    title = "A Dataset of Peer Reviews ({P}eer{R}ead): Collection, Insights and {NLP} Applications",
    author = "Kang, Dongyeop  and
      Ammar, Waleed  and
      Dalvi, Bhavana  and
      van Zuylen, Madeleine  and
      Kohlmeier, Sebastian  and
      Hovy, Eduard  and
      Schwartz, Roy",
    editor = "Walker, Marilyn  and
      Ji, Heng  and
      Stent, Amanda",
    booktitle = "Proceedings of the 2018 Conference of the North {A}merican Chapter of the Association for Computational Linguistics: Human Language Technologies, Volume 1 (Long Papers)",
    month = jun,
    year = "2018",
    address = "New Orleans, Louisiana",
    publisher = "Association for Computational Linguistics",
    url = "https://aclanthology.org/N18-1149/",
    doi = "10.18653/v1/N18-1149",
    pages = "1647--1661"
}

@inproceedings{DBLP:conf/nips/StelmakhSS19,
  author       = {Ivan Stelmakh and
                  Nihar B. Shah and
                  Aarti Singh},
  editor       = {Hanna M. Wallach and
                  Hugo Larochelle and
                  Alina Beygelzimer and
                  Florence d'Alch{\'{e}}{-}Buc and
                  Emily B. Fox and
                  Roman Garnett},
  title        = {On Testing for Biases in Peer Review},
  booktitle    = {Advances in Neural Information Processing Systems 32: Annual Conference
                  on Neural Information Processing Systems 2019, NeurIPS 2019, December
                  8-14, 2019, Vancouver, BC, Canada},
  pages        = {5287--5297},
  year         = {2019},
  url          = {https://proceedings.neurips.cc/paper/2019/hash/d3d80b656929a5bc0fa34381bf42fbdd-Abstract.html},
  bibsource    = {dblp computer science bibliography, https://dblp.org}
}

@article{DBLP:journals/jmlr/ShahTMGL18,
  author       = {Nihar B. Shah and
                  Behzad Tabibian and
                  Krikamol Muandet and
                  Isabelle Guyon and
                  Ulrike von Luxburg},
  title        = {Design and Analysis of the {NIPS} 2016 Review Process},
  journal      = {J. Mach. Learn. Res.},
  volume       = {19},
  pages        = {49:1--49:34},
  year         = {2018},
  url          = {https://jmlr.org/papers/v19/17-511.html},
  bibsource    = {dblp computer science bibliography, https://dblp.org}
}

@inproceedings{DBLP:conf/emnlp/BowmanAPM15,
  author       = {Samuel R. Bowman and
                  Gabor Angeli and
                  Christopher Potts and
                  Christopher D. Manning},
  editor       = {Llu{\'{\i}}s M{\`{a}}rquez and
                  Chris Callison{-}Burch and
                  Jian Su and
                  Daniele Pighin and
                  Yuval Marton},
  title        = {A large annotated corpus for learning natural language inference},
  booktitle    = {Proceedings of the 2015 Conference on Empirical Methods in Natural
                  Language Processing, {EMNLP} 2015, Lisbon, Portugal, September 17-21,
                  2015},
  pages        = {632--642},
  publisher    = {The Association for Computational Linguistics},
  year         = {2015},
  url          = {https://doi.org/10.18653/v1/d15-1075},
  doi          = {10.18653/V1/D15-1075},
  bibsource    = {dblp computer science bibliography, https://dblp.org}
}

@inproceedings{DBLP:conf/naacl/WilliamsNB18,
  author       = {Adina Williams and
                  Nikita Nangia and
                  Samuel R. Bowman},
  editor       = {Marilyn A. Walker and
                  Heng Ji and
                  Amanda Stent},
  title        = {A Broad-Coverage Challenge Corpus for Sentence Understanding through
                  Inference},
  booktitle    = {Proceedings of the 2018 Conference of the North American Chapter of
                  the Association for Computational Linguistics: Human Language Technologies,
                  {NAACL-HLT} 2018, New Orleans, Louisiana, USA, June 1-6, 2018, Volume
                  1 (Long Papers)},
  pages        = {1112--1122},
  publisher    = {Association for Computational Linguistics},
  year         = {2018},
  url          = {https://doi.org/10.18653/v1/n18-1101},
  doi          = {10.18653/V1/N18-1101},
  bibsource    = {dblp computer science bibliography, https://dblp.org}
}

@inproceedings{DBLP:conf/naacl/DevlinCLT19,
  author       = {Jacob Devlin and
                  Ming{-}Wei Chang and
                  Kenton Lee and
                  Kristina Toutanova},
  editor       = {Jill Burstein and
                  Christy Doran and
                  Thamar Solorio},
  title        = {{BERT:} Pre-training of Deep Bidirectional Transformers for Language
                  Understanding},
  booktitle    = {Proceedings of the 2019 Conference of the North American Chapter of
                  the Association for Computational Linguistics: Human Language Technologies,
                  {NAACL-HLT} 2019, Minneapolis, MN, USA, June 2-7, 2019, Volume 1 (Long
                  and Short Papers)},
  pages        = {4171--4186},
  publisher    = {Association for Computational Linguistics},
  year         = {2019},
  url          = {https://doi.org/10.18653/v1/n19-1423},
  doi          = {10.18653/V1/N19-1423},
  bibsource    = {dblp computer science bibliography, https://dblp.org}
}

@article{DBLP:journals/corr/abs-1907-11692,
  author       = {Yinhan Liu and
                  Myle Ott and
                  Naman Goyal and
                  Jingfei Du and
                  Mandar Joshi and
                  Danqi Chen and
                  Omer Levy and
                  Mike Lewis and
                  Luke Zettlemoyer and
                  Veselin Stoyanov},
  title        = {RoBERTa: {A} Robustly Optimized {BERT} Pretraining Approach},
  journal      = {CoRR},
  volume       = {abs/1907.11692},
  year         = {2019},
  url          = {http://arxiv.org/abs/1907.11692},
  eprinttype    = {arXiv},
  eprint       = {1907.11692},
  bibsource    = {dblp computer science bibliography, https://dblp.org}
}

@inproceedings{DBLP:conf/acl/NieWDBWK20,
  author       = {Yixin Nie and
                  Adina Williams and
                  Emily Dinan and
                  Mohit Bansal and
                  Jason Weston and
                  Douwe Kiela},
  editor       = {Dan Jurafsky and
                  Joyce Chai and
                  Natalie Schluter and
                  Joel R. Tetreault},
  title        = {Adversarial {NLI:} {A} New Benchmark for Natural Language Understanding},
  booktitle    = {Proceedings of the 58th Annual Meeting of the Association for Computational
                  Linguistics, {ACL} 2020, Online, July 5-10, 2020},
  pages        = {4885--4901},
  publisher    = {Association for Computational Linguistics},
  year         = {2020},
  url          = {https://doi.org/10.18653/v1/2020.acl-main.441},
  doi          = {10.18653/V1/2020.ACL-MAIN.441},
  bibsource    = {dblp computer science bibliography, https://dblp.org}
}

@inproceedings{DBLP:conf/acl/GururanganMSLBD20,
  author       = {Suchin Gururangan and
                  Ana Marasovic and
                  Swabha Swayamdipta and
                  Kyle Lo and
                  Iz Beltagy and
                  Doug Downey and
                  Noah A. Smith},
  editor       = {Dan Jurafsky and
                  Joyce Chai and
                  Natalie Schluter and
                  Joel R. Tetreault},
  title        = {Don't Stop Pretraining: Adapt Language Models to Domains and Tasks},
  booktitle    = {Proceedings of the 58th Annual Meeting of the Association for Computational
                  Linguistics, {ACL} 2020, Online, July 5-10, 2020},
  pages        = {8342--8360},
  publisher    = {Association for Computational Linguistics},
  year         = {2020},
  url          = {https://doi.org/10.18653/v1/2020.acl-main.740},
  doi          = {10.18653/V1/2020.ACL-MAIN.740},
  bibsource    = {dblp computer science bibliography, https://dblp.org}
}

@article{anil2023gemini,
  title     = {Gemini: A Family of Highly Capable Multimodal Models},
  author    = {Anil, Rohan and Borgeaud, Sebastian and Alayrac, Jean-Baptiste and Yu, Jiahui and Soricut, Radu and Schalkwyk, Johan and Dai, Andrew M. and Hauth, Anja and Millican, Katie and Silver, David and others},
  journal   = {arXiv preprint arXiv:2312.11805},
  year      = {2023},
  note      = {Version v5}
}

@misc{meta2024llama4,
  author       = {{Meta AI}},
  title        = {{LLaMA 4: Multimodal Intelligence}},
  year         = {2024},
  howpublished = {\url{https://ai.meta.com/blog/llama-4-multimodal-intelligence/}},
  note         = {Accessed March 2025}
}

@article{grattafiori2024llama3,
  title     = {The Llama 3 Herd of Models},
  author    = {Grattafiori, Aaron and Dubey, Abhimanyu and Jauhri, Abhinav and Pandey, Abhinav and Kadian, Abhishek and Al-Dahle, Ahmad and others},
  journal   = {arXiv preprint arXiv:2407.21783},
  year      = {2024},
  doi       = {10.48550/arXiv.2407.21783},
  note      = {Version v3}
}

@article{yang2025qwen3,
  title     = {Qwen3 Technical Report},
  author    = {Yang, An and Li, Anfeng and Yang, Baosong and Zhang, Beichen and Hui, Binyuan and Zheng, Bo and Yu, Bowen and Gao, Chang and Huang, Chengen and Lv, Chenxu and Zheng, Chujie and Liu, Dayiheng and Zhou, Fan and Huang, Fei and Hu, Feng and Ge, Hao and Wei, Haoran and Lin, Huan and Tang, Jialong and Yang, Jian and Tu, Jianhong and Zhang, Jianwei and Yang, Jianxin and Yang, Jiaxi and Zhou, Jing and Zhou, Jingren and Lin, Junyang and Dang, Kai and Bao, Keqin and Yang, Kexin and Yu, Le and Deng, Lianghao and Li, Mei and Xue, Mingfeng and Li, Mingze and Zhang, Pei and Wang, Peng and Zhu, Qin and Men, Rui and Gao, Ruize and Liu, Shixuan and Luo, Shuang and Li, Tianhao and Tang, Tianyi and Yin, Wenbiao and Ren, Xingzhang and Wang, Xinyu and Zhang, Xinyu and Ren, Xuancheng and Fan, Yang and Su, Yang and Zhang, Yichang and Zhang, Yinger and Wan, Yu and Liu, Yuqiong and Wang, Zekun and Cui, Zeyu and Zhang, Zhenru and Zhou, Zhipeng and Qiu, Zihan},
  journal   = {arXiv preprint arXiv:2505.09388},
  year      = {2025},
  doi       = {10.48550/arXiv.2505.09388}
}

@article{openai2024gpt4,
  title     = {GPT-4 Technical Report},
  author    = {{OpenAI}},
  journal   = {arXiv preprint arXiv:2303.08774},
  year      = {2024},
  url       = {https://arxiv.org/abs/2303.08774}
}

@article{DBLP:journals/corr/abs-2105-03011,
  author       = {Pradeep Dasigi and
                  Kyle Lo and
                  Iz Beltagy and
                  Arman Cohan and
                  Noah A. Smith and
                  Matt Gardner},
  title        = {A Dataset of Information-Seeking Questions and Answers Anchored in
                  Research Papers},
  journal      = {CoRR},
  volume       = {abs/2105.03011},
  year         = {2021},
  url          = {https://arxiv.org/abs/2105.03011},
  eprinttype    = {arXiv},
  eprint       = {2105.03011},
  bibsource    = {dblp computer science bibliography, https://dblp.org}
}

\appendix
\section*{Appendix}

\section{Additional Implementation Details}
\label{app:impl}

For Gemini-based models, the maximum number of generated tokens is fixed to 4096, while OpenAI and LLaMA models have a maximum output of 8,192 new tokens.

\textsc{IMPACT-P} uses proprietary models for deliberation and adjudication, with LLaMA-4-Maverick and Gemini-3-flash serving as Deliberative Intensity Agents (DIA), GPT-5.2 as the adjudication agent, and ACEA providing aspect-conditioned evidence selection. \textsc{IMPACT-OA} replaces proprietary components with openly accessible models, using Qwen3-32b and GPT-OSS-20b as DIA agents and LLaMA-4-Maverick as the adjudication agent, while using LLaMA-3.3-70b for ACEA. We discuss the role and prompt used by ACEA, DIAs, and the adjudication agent in detail in Appendix~\ref{Appendix: Judge} and ~\ref{Appendix: case_study}. 
All generic multi-agent baselines (e.g., Debate, MAD, Self-Refine, ChatEval, CourtEval) were implemented under a controlled setting to ensure fair comparison. We used the same underlying LLM (GPT-5.2) and consistent prompting across all baselines. For iterative frameworks, we followed the termination criteria specified in their respective original papers. The per-device batch size is set to 2 with gradient accumulation over 8 steps (effective batch size of 16). We use a 20\% validation split and an 80\% test split, with a fixed random seed of 42.

LoRA adapters are applied to the attention and feed-forward projection layers with rank 8, scaling factor 16, and dropout 0.1. Training is conducted in \texttt{bfloat16} precision with a maximum sequence length of 4096 tokens, and loss is computed only on assistant responses. The best checkpoint is selected based on validation loss. During both training and inference, the LLM has a maximum output of 768 new tokens. During inference, the fine-tuned model is evaluated using beam search with 2 beams and a temperature of 0.01. All training and inference experiments are conducted on NVIDIA A100 GPUs with 80~GB memory, using two compute nodes.

\section{Token Efficiency Analysis}

We randomly sampled 100 instances and computed the average number of tokens used by each method. The results are shown in Table~\ref{tab:token_cost}. We use a standard GPT tokenizer\footnote{\url{https://platform.openai.com/tokenizer}} for this calculation. These results show that \textsc{TIDE} is substantially more token-efficient than multi-agent baselines.

\begin{table}[!ht]
\centering
\small
\begin{tabular}{lc}
\toprule
Method & Avg Tokens per Sample \\
\midrule
TIDE & 1892.9 \\
CoT (Single) & 2410.1 \\
IMPACT-P & 36184.9 \\
IMPACT-OA & 39412.6 \\
Self-Refine & 18741.5 \\
Debate & 26483.8 \\
MAD & 23861.4 \\
ChatEval & 34712.9 \\
CourtEval & 37184.2 \\
\bottomrule
\end{tabular}
\caption{Average token usage per sample across methods.}
\label{tab:token_cost}
\end{table}

\section{Result: FNR and FPR value}
\label{Appendix:fnr}

    \begin{figure}[ht]
      \fbox{%
        \includegraphics[width=\columnwidth]{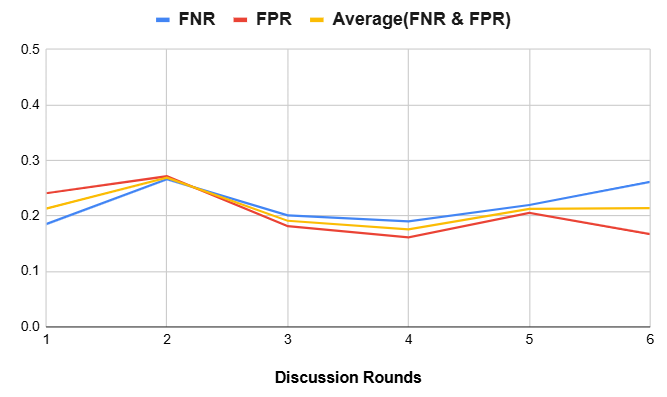}
      }
      \caption{False Negative Rate (FNR), False Positive Rate (FPR), and their average as a function of the number of discussion rounds in the multi-agent deliberation framework.}
      \label{fig:experiments2}
    \end{figure}

Across multiple experimental runs in which the number of discussion rounds in the multi-agent deliberation framework is varied as a hyperparameter, we further analyze error characteristics of the framework in detecting contradictions by examining the false negative rate (FNR), false positive rate (FPR), and their average (see Figure \ref{fig:experiments2}). With a single discussion round, the average error rate is 0.2131. Increasing the discussion depth to two rounds leads to a sharp increase in the average error rate to 0.2689 (+26.1\%), indicating that limited deliberation can amplify inconsistent or overconfident judgments. However, as the number of rounds increases beyond this point, error rates decrease substantially: the average drops to 0.1914 (-28.82\%) at three rounds and reaches its minimum at four rounds (0.1757, -8.2\%), corresponding to simultaneous reductions in both FNR and FPR. Further increases in discussion depth yield diminishing and unstable effects, with the average error rising again to 0.2127 at five rounds and 0.2143 at six rounds, driven by renewed increases in false negatives and fluctuations in false positives. Consistent with the agreement analysis, these results indicate that four discussion rounds achieve the best balance between missed and spurious detections.

\section{Annotation Details} \label{Appendix: annotation_details}

\begin{figure}[t]
\centering
\begin{tikzpicture}
\begin{axis}[
    ybar,
    bar width=10pt,
    ymin=0,
    ymax=25,
    ylabel={Percentage (\%)},
    symbolic x coords={SOU,SUB,ORG,MOT,CMP,CLR},
    xtick=data,
    x tick label style={rotate=25, anchor=east},
    nodes near coords,
    nodes near coords align={vertical},
    width=\linewidth,
    height=5cm
]
\addplot coordinates {
    (SOU,19.45)
    (SUB,17.86)
    (ORG,15.68)
    (MOT,14.19)
    (CMP,11.55)
    (CLR,21.28)
};
\end{axis}
\end{tikzpicture}
\caption{Percentage distribution of contradictions across aspects in the human-annotated test dataset.
SOU: Soundness, SUB: Substance, ORG: Originality, MOT: Motivation, CMP: Meaningful Comparison, CLR: Clarity.}
\label{fig:aspect_dist_training}
\end{figure}
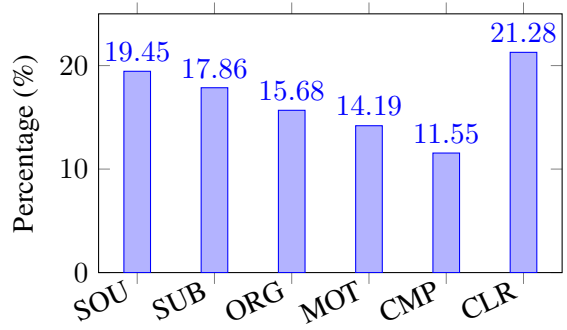

\begin{figure}[t]
\centering
\begin{tikzpicture}
\begin{axis}[
    ybar,
    bar width=18pt,
    ymin=0,
    ymax=50,
    ylabel={Percentage (\%)},
    xlabel={Intensity Level},
    symbolic x coords={1,2,3},
    xtick=data,
    nodes near coords,
    nodes near coords align={vertical},
    width=0.75\linewidth,
    height=7cm
]
\addplot coordinates {
    (1,26)
    (2,45.3)
    (3,28.7)
};
\end{axis}
\end{tikzpicture}
\caption{Percentage distribution of contradiction intensities in the human-annotated test dataset.}
\label{fig:intensity_dist_training}
\end{figure}
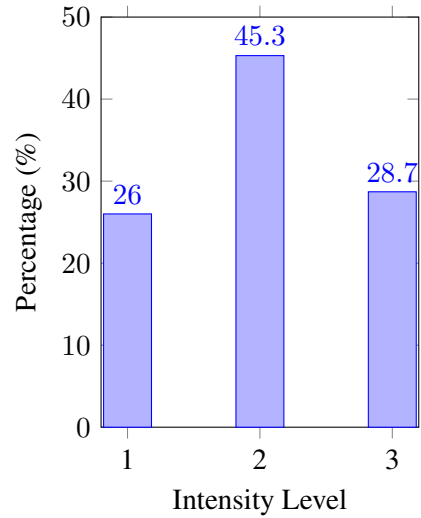

\begin{figure}[t]
\centering
\begin{tikzpicture}
\begin{axis}[
    ybar,
    bar width=10pt,
    ymin=0,
    ymax=25,
    ylabel={Percentage (\%)},
    symbolic x coords={SOU,SUB,ORG,MOT,CMP,CLR},
    xtick=data,
    x tick label style={rotate=25, anchor=east},
    nodes near coords,
    nodes near coords align={vertical},
    width=\linewidth,
    height=5cm
]
\addplot coordinates {
    (SOU,14.0)
    (SUB,18.7)
    (ORG,15.7)
    (MOT,20.5)
    (CMP,10.7)
    (CLR,20.2)
};
\end{axis}
\end{tikzpicture}
\caption{Aspect-wise percentage distribution of contradictions in synthetic data. 
SOU: Soundness, SUB: Substance, ORG: Originality, MOT: Motivation, CMP: Meaningful Comparison, CLR: Clarity.}
\label{fig:aspect_dist}
\end{figure}
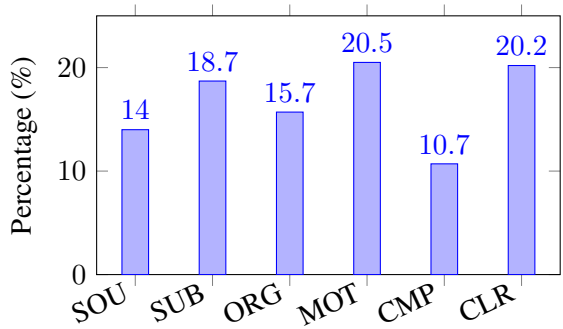

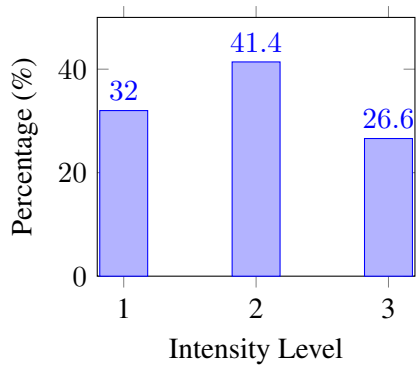
\begin{figure}[t]
\centering
\begin{tikzpicture}
\begin{axis}[
    ybar,
    bar width=18pt,
    ymin=0,
    ymax=50,
    ylabel={Percentage (\%)},
    xlabel={Intensity Level},
    symbolic x coords={1,2,3},
    xtick=data,
    nodes near coords,
    nodes near coords align={vertical},
    width=0.75\linewidth,
    height=5cm
]
\addplot coordinates {
    (1,32.0)
    (2,41.4)
    (3,26.6)
};
\end{axis}
\end{tikzpicture}
\caption{Intensity-wise percentage distribution of contradictions present in synthetic data.}
\label{fig:intensity_dist}
\end{figure}

\subsection{Annotation Guidelines} \label{sec:guidelines}

The goal of the annotation task is to identify contradictions between two reviews written for the same paper, determine the severity of the contradiction, and give a brief explanation on the severity of the score given. Follow the steps below when annotating any pair of reviews.

\textbf{Step 1: Understand the paper.}  
Read the \textbf{title} and \textbf{abstract} of the paper to understand the background, motivation, and main contributions made by the paper. You may read these sections multiple times for better clarity. If a reviewer comment refers to specific technical details, you may briefly consult the paper to clarify the context.

\textbf{Step 2: Read the reviews carefully.}  
Read both reviews thoroughly. Pay attention to the opinions and judgments expressed by each reviewer, identify the aspects of the paper being discussed and the reviewers' sentiment behind the same (e.g., novelty, soundness, clarity, experiments, motivation). Do not treat individual sentences in isolation; consider the full context of each review.

\textbf{Step 3: Identify contradictions.}  
Identify pairs of statements that refer to the same aspect of the paper and determine whether they contradict each other. If multiple sentences contribute to a contradiction, then all of them must be included as evidences. A contradiction exists when the two comments express mutually incompatible judgments about the same issue. The contradiction could be explicit as well as implicit. The statements must have opposing sentiments (eg. positive vs. negative) in order for them to be considered contradictory. 

To decide whether a contradiction is present, ask the following question:

\begin{quote}
``Would it be very unlikely that both of these statements could be true at the same time?''
\end{quote}

If the answer is yes, treat the pair as a contradiction. Differences in emphasis, suggestions for improvement, or unrelated critiques should not be marked as contradictions.

\textbf{Step 4: Write the contradiction statement.}  
For each contradiction, write a \textbf{contradiction statement} explaining why the two reviews disagree. The statement should clearly describe the shared aspect and summarize the opposing viewpoints in a neutral manner, without copying text from the reviews.

\textbf{Step 5: Assign contradiction severity.}  
Assign a severity score to each contradiction using the rubric below.

\begin{itemize}
    \item \textbf{Score 1 --- Low Severity (Implicit Contradiction):} One statement is generic while the other is specific; the conflict is indirect or interpretative; the disagreement is weak or implicit; there is no strong positive or negative polarity. Eg.
    \begin{quote}
"review1": "Section 3, where the authors describe the proposed techniques is somewhat confusing to read, because of a lack of detailed mathematical explanations of the proposed techniques."

\end{quote}
\begin{quote}
   "review2": "The paper is clearly written and the results seem compelling." 
\end{quote}

    \item \textbf{Score 2 --- Moderate Severity (Explicit but Mild Conflict):} Both statements refer to the same aspect; one statement is mildly critical while the other is moderately or strongly positive; the disagreement is clear but not extreme.
    \begin{quote}
        "review1": "However, the novelty is limited in the sense it is application of coordinate descent on power iterations."
    \end{quote}
    \begin{quote}
    "review2": "This paper appears to be the first to solve this problem, and make a connection to coordinate decent."
    \end{quote}
    \item \textbf{Score 3 --- High Severity (Direct Strong Contradiction):} One statement is strongly positive while the other is strongly negative; the judgments are highly polarized; the conflict is clear, direct, and fundamental.
    \begin{quote}
    "The paper is clearly written."
    \end{quote}
    \begin{quote}
    "I found the presentation of the proposed measure overly confusing."
    \end{quote}
    
\end{itemize}

You must base the intensity judgment on the strength and polarity of the evaluative language rather than the length or technical details of the comments.

\textbf{Additional Notes.}  
A contradiction may still exist when one review provides a general evaluation and the other comments on a specific part of the paper. Focus only on the given aspect category when comparing comments, and do not introduce personal opinions or external knowledge during annotation.



\subsection{Annotator Training and Annotation Process}
Given the technical complexity of scientific peer reviews and their frequent use of domain-specific language, we recruited six doctoral students (4th--5th year) in Machine Learning as annotators, each with several years of experience in conducting research, publishing in peer-reviewed venues, and participating in the peer-review process. To initialize the annotation process, two domain experts with more than ten years of experience in scientific publishing independently annotated an initial set of 75 review pairs. The experts subsequently met to discuss discrepancies and reconcile their annotations, resulting in a set of expert-validated reference annotations. These reference annotations were used to support a multi-phase annotator training procedure. In the first phase, each annotator independently annotated 25 review pairs. Their annotations were then reviewed jointly with the experts to identify systematic errors in contradiction evidence selection, intensity interpretation, and guideline adherence. Based on this analysis, we refined the annotation guidelines and provided targeted feedback to the annotators. Two additional training phases followed, each involving newly sampled review pairs, enabling annotators to iteratively improve their understanding of both evidence-level contradiction identification and graded intensity assignment. From the third training round onward, most annotators exhibited stable proficiency, correctly identifying expert-aligned contradiction evidence spans and assigning matching intensity labels for at least 70\% of expert-annotated contradiction instances, where correctness requires agreement with experts on both evidence selection and intensity assignment.

Throughout the full annotation process, we continuously monitored the annotated data to detect inconsistencies, ambiguous cases, and sources of systematic confusion. We employed an iterative feedback mechanism in which annotators received regular guidance informed by expert review and error analysis. In cases of disagreement or uncertainty, domain experts adjudicated and provided final decisions, ensuring consistency and coherence across the dataset. Following the completion of annotation, we assessed inter-annotator agreement separately for the two core components of the task. Agreement on contradiction evidence identification was evaluated based on semantic alignment with expert-validated evidence spans, while agreement on contradiction intensity labels was measured using Cohen’s kappa \cite{cohen1960coefficient}. We obtained an average kappa score of 0.64 for intensity annotation, indicating substantial agreement given the fine-grained, evidence-level nature of the task. Overall, these results demonstrate the effectiveness of the training protocol and annotation design, while reflecting the inherent difficulty of jointly annotating explicit contradiction evidence and the graded intensity of disagreement in scientific peer reviews.

\subsection{Subjectivity and Borderline Cases in Intensity Annotation}
\label{sec:intensity_subjectivity}

Contradiction intensity annotation is inherently subjective, particularly when reviewer comments involve hedging language (e.g., ``seems unclear,'' ``might be incremental'') or when reviewers weigh trade-offs differently. Although we define a formal scoring rubric (Appendix~\ref{sec:annotation_guidelines}), some cases remain difficult to categorize consistently across annotators.

We emphasize that RevCI annotates contradiction evidence pairing and graded contradiction intensity, while aspect categories and aspect spans are inherited from the ASAP annotations and were not re-annotated in this work. Agreement was measured on whether a span constitutes a contradiction and its corresponding intensity label.

We measure agreement using Cohen’s $\kappa$ over the full 4-class label space $\{0,1,2,3\}$, yielding an overall $\kappa = 0.64$, indicating substantial agreement given the fine-grained, evidence-level nature of the task. To provide label-level granularity, we additionally compute one-vs-rest $\kappa$ for each class: $\kappa_0 = 0.81$, $\kappa_1 = 0.47$, $\kappa_2 = 0.63$, and $\kappa_3 = 0.72$.

As expected for an ordinal severity scale, agreement is highest for null (0) and strong (3) contradictions and lowest for mild contradictions (1), which often involve hedged or scalar evaluative language. Most residual disagreements occur between adjacent levels ($1 \leftrightarrow 2$, $2 \leftrightarrow 3$), with rare extreme confusions ($1 \leftrightarrow 3$). Consistent with this ordinal structure, weighted $\kappa \approx 0.70$, indicating that disagreements primarily reflect calibration differences rather than categorical inconsistencies.

We present representative borderline cases below where contradiction intensity was difficult to define.

\paragraph{Case 1: Apparent doubt vs. non-contradiction.}
A few borderline cases emerged during annotation where apparent doubt between reviews could be misinterpreted as contradiction when no true contradiction exists. For instance, Reviewer~1 stated:
\begin{quote}
``In the image-matching experiment, is it possible to add results for an LSSVM or other baseline besides [9]?''
\end{quote}
while Reviewer~2 wrote:
\begin{quote}
``Experiments show that the proposed method is as accurate but $>{}10\times$ faster than traditional large-margin learning techniques on synthetic data and an image alignment problem.''
\end{quote}
Although both comments concern experimental evaluation, the first reviewer merely requests additional baselines without disputing the reported results. Some annotators initially perceived this as a low-conflict contradiction because one review sounds more cautious while the other is strongly positive. However, based on our annotation definition, the pair was removed since opposing doubt cannot be counted as a contradiction.

\paragraph{Case 2: Distinguishing between Scores 0 and 1.}
More challenging distinctions appeared between Scores~0 and~1. In one case, Reviewer~1 wrote:
\begin{quote}
``The evaluation is a good start with comparing several base DA methods with and without the proposed TransferNorm architecture,''
\end{quote}
while Reviewer~2 stated:
\begin{quote}
``The experiments are extensively evaluated both qualitatively and quantitatively, demonstrating the effectiveness of the proposed TranNorm.''
\end{quote}
Here, both reviewers explicitly judge evaluation thoroughness, and the polarity difference is subtle. Some annotators initially leaned toward Score~0 because both claims appear positive. However, the first reviewer’s phrasing (``good start'') is hedged and implies that further work is needed rather than expressing a strongly positive judgment. Consequently, the pair was labeled Score~1.

\paragraph{Case 3: Distinguishing between Scores 1 and 2.}
A few distinctions also appeared between Scores~1 and~2. In one case, Reviewer~1 noted:
\begin{quote}
``I suspect that most of the information is stored in the memory and only a small change of the training data is allowed... the high Inception Score cannot show the generalization ability as well...''
\end{quote}
raising explicit concerns about generalization and the adequacy of the reported metric. Reviewer~2, however, stated:
\begin{quote}
``Overall, I think MemoryGAN opened a new direction of GAN and is worth further exploration.''
\end{quote}
Here, the reviewers express opposing evaluative stances toward the method’s quality and promise. Annotators initially considered Score~1 because the second comment is relatively generic. However, the positive judgment directly contrasts with the first reviewer’s technical skepticism regarding soundness. Consequently, the pair was labeled Score~2, reflecting an explicit but moderate disagreement.
\subsection{Human Annotation Protocol} \label{sec:human_annotation}
We first filtered review pairs to match the annotators’ domain expertise and prior reviewing experience, and then randomly sampled 800 pairs from this pool for annotation. For each selected pair, annotators read both reviews in full, identify explicit evidence spans from each review that express contradictory judgments, and write a contradiction explanation (CE) explaining the disagreement. Contradiction intensity is annotated at the evidence level, allowing a single review pair to contain multiple contradictions with varying strengths. Annotators then assign a contradiction intensity score in $\{1,2,3\}$ to each evidence pair using a 3-point Likert scale, where \textbf{1} denotes \emph{low severity} (weak or implicit contradictions), \textbf{2} denotes \emph{moderate severity} (explicit but mild contradictions), and \textbf{3} denotes \emph{high severity} (direct and strong contradictions). The rubric and detailed annotation guidelines are described in Section~\ref{sec:guidelines}. Requiring explicit evidence spans together with contradiction explanation ensures the faithfulness of the annotations and encourages careful reasoning about both the source and severity of disagreement, resulting in higher-quality annotations.

\subsection{Annotator Compensation}
Annotators were compensated at an hourly rate consistent with standard doctoral research assistant wages, in accordance with institutional guidelines. Compensation was based on time spent on annotation rather than on a per-instance basis, reflecting the substantial variability in review length, technical complexity, and cognitive effort required to identify contradiction evidence, assign graded intensity labels, and produce faithful explanations. To support annotation quality and reduce fatigue, we imposed a maximum daily annotation limit of six hours per annotator. This constraint ensured that annotators had sufficient time to carefully read full reviews, reason about conflicting judgments, and apply annotation guidelines consistently. On average, annotating one review pair took approximately 30--40 minutes, depending on review length and technical complexity. All compensation practices complied with institutional policies and applicable ethical standards.

\section{Evaluation Metrics}
\label{Appendix:metrics}
Contradiction evidence extraction yields unordered sets of instances with variable cardinality and variable-length evidence, making count-based evaluation inadequate.
For example, a ground-truth evidence span may be \textit{``The authors claim a significant improvement in accuracy, but the reported results are not statistically significant,''} while a predicted evidence may be \textit{``the reported results are not statistically significant.''} In such cases, exact-match evaluation fails, and token-level F1 is heavily penalized due to missing tokens, whereas ROUGE-L assigns a high score by capturing the longest common subsequence.

Given the set of ground-truth contradictions $G$ and predicted contradictions $P$, we enforce a one-to-one alignment by computing the maximum-weight matching using the Hungarian algorithm as implemented in \texttt{scipy.optimize.linear\_sum\_assignment}~\cite{scipy} at the sentence level, and averaging the aligned ROUGE-L scores. This prevents a single prediction from being aligned with multiple ground-truth instances (or vice versa) and assigns equal weight to both sentences in each evidence pair. When $|G| \neq |P|$, global matching may result in weak alignments. To mitigate this, we discard aligned pairs whose ROUGE-L score falls below a threshold $\lambda_{\text{match}}$\footnote{We set the value of $\lambda_{\text{match}} = 0.3$ empirically.}. Discarded pairs, along with other unmatched instances, are treated as unmatched and contribute to evidence-level false negatives (FN) and false positives (FP).

Finally, to evaluate the model’s ability to predict evidence intensity, we compute agreement metrics over the matched evidence pairs. Specifically, we report Cohen’s Kappa ($\kappa$)~\cite{cohen1960coefficient}, Spearman’s rank correlation ($\rho$)~\cite{spearman1961proof}, and Kendall’s Tau ($\tau$)~\cite{kendall1938new}.

\section{Error Analysis} \label{Appendix_error}

We perform qualitative evaluation to understand where our proposed model fails.

\paragraph{Ambiguity in Evaluative Language:}
A common source of error arises from vague or scalar evaluative terms, such as “good start,” “extensive,” or “reasonable.” In several cases, differences in emphasis for example, one reviewer describing the evaluation as “a good start” while suggesting additional comparisons and another characterizing the experiments as “extensively evaluated” do not constitute genuine contradictions but rather reflect differing thresholds or expectations. However, the framework sometimes interprets such complementary or gradational assessments as contradictory, leading to false positives.

\paragraph{Aspect Confusion:}
A recurring source of error stems from confusion in aspect identification, particularly when a single sentence references multiple aspects simultaneously (e.g., novelty, evaluation, and clarity). In such cases, the model frequently misclassifies the primary aspect of disagreement by anchoring on salient lexical cues rather than the reviewer’s central evaluative intent. As a result, contradictions are often assigned to an incorrect aspect (e.g., Soundness instead of Originality).

\section{LLM Prompts for Evaluation} \label{app:llm_prompts}

\begin{promptbox}{Prompt For Evaluating Baseline LLM Models} 

Carefully analyze the two peer reviews provided. Your objective is to identify contradictions between the reviews based on the defined contradiction criteria and within the specified evaluation aspects.

Contradiction Definition: \{contradiction\_definition\}

Follow the steps below:

Analyze the two peer reviews carefully.

Identify contradictions between them in accordance with the provided contradiction definition and the specified aspects.

Support each identified contradiction with explicit statements from both reviews.

Assign each contradiction to exactly one of the given aspects.

Ensure that no two contradictions rely on the same pair of evidence sentences.

Assign an intensity score for each contradiction based on the provided scoring metric.

\{Aspect names\}: \{Aspect Definition\}

\{Scoring Metric\}:

\{Example outputs\}:

\{Output Format\}:

\{Key Rules of Contradiction Extraction\}:

\end{promptbox}

This appendix lists the exact prompts used to evaluate all baseline LLMs and \textsc{IMPACT} agents under a unified protocol.

\begin{promptbox}{ACEA PROMPT} \label{Appendix: Judge}
\textbf{CONTRADICTION DEFINITION:}
A contradiction occurs when two statements make claims that cannot both be true simultaneously. Contradictions can be implicit as well as explicit.

\textbf{TASK}:
Analyze these two peer reviews and identify both implicit as well as explicit contradictions specifically related to the "\{aspect\_name\}" aspect.

\textbf{ASPECT FOCUS}: 
\{aspect\_name\}
Description: \{aspect\_description\}

\textbf{REVIEW 1}:
\{review\_1\}
\textbf{REVIEW 2}:
\{review\_2\}

\textbf{INSTRUCTIONS}:
1. Look for statements in both reviews that relate to "
\{aspect\_name\}"
2. Identify if these statements contradict each other
3. If contradictions exist, extract them with exact evidence from both reviews
4. Extract as many contradictions you can find related to "\{aspect\_name\}"
5. It is NOT necessary that you find contradictions for this Aspect; if none exist, return an empty list

\textbf{Output format}:
\{\{
  "aspect": "\{aspect\_name\}",
  "contradictions": (
    \{\{
      "contradiction": "brief description of contradiction",
      "evidence": ("exact quote from Review 1", "exact quote from Review 2")
    \}\}
  )
\}\}

\textbf{RULES}:
- Focus ONLY on the "{aspect\_name}" aspect
- Evidence array must have exactly 2 elements: [Review\_1\_quote, Review\_2\_quote]
- Extract the exact complete sentences from each review that illustrate the contradiction
- If no contradictions found for this aspect, return empty contradictions array
- Output ONLY valid JSON, no additional text
\end{promptbox}
\begin{promptbox}{DIA (Initial Scoring)}

You are an intensity scorer for peer-review POTENTIAL contradictions.

\textbf{Scoring Rubric}:
- Score 0 — No Contradiction (Compatible or Orthogonal Statements)
Statements refer to different aspects, topics, or evaluation criteria, OR
Statements discuss the same aspect but are fully compatible or consistent, OR
Any differences are descriptive, complementary, or additive rather than conflicting.
EXAMPLES:- \{2 examples\}

- Score 1 — Low Severity (Implicit Contradiction) One statement is generic, the other is specific, OR The conflict is indirect or interpretative, No strong positive/negative polarity, Weak or implicit disagreement.
EXAMPLES:- \{2 examples\}

- Score 2 — Moderate Severity (Explicit but Mild Conflict) Both statements explicitly refer to the same aspect, One gives light criticism, the other is mildly or significantly positive, Explicit but not extreme polarity.
EXAMPLES:- \{2 examples\}

- Score 3 — High Severity (Direct Strong Contradiction) Strongly worded positive vs. negative evaluation of the same aspect, Extremely polarized opposite judgments, Clear and fundamental extreme disagreement.
EXAMPLES:- \{2 examples\}

\textbf{REVIEW 1 CONTEXT}:
\{review\_1\}

\textbf{REVIEW 2 CONTEXT}:
\{review\_2\}

\textbf{EVIDENCE FROM REVIEW 1}:
\{s1\}

\textbf{EVIDENCE FROM REVIEW 2}:
\{s2\}

Analyze these two pieces of evidence and determine the intensity of contradiction.

\textbf{Output format}:
\{\{
  "intensity": <0, 1, 2, or 3>,
  "reasoning": "detailed explanation of why you assigned this intensity score"
\}\}

\end{promptbox}
\begin{promptbox}{DIA (Debate Prompt)}

You are an intensity scorer for peer-review POTENTIAL contradictions.

\textbf{Scoring Rubric}:
- Score 0 — No Contradiction (Compatible or Orthogonal Statements)
Statements refer to different aspects, topics, or evaluation criteria, OR
Statements discuss the same aspect but are fully compatible or consistent, OR
Any differences are descriptive, complementary, or additive rather than conflicting.
EXAMPLES:- \{2 examples\}

- Score 1 — Low Severity (Implicit Contradiction) One statement is generic, the other is specific, OR The conflict is indirect or interpretative, No strong positive/negative polarity, Weak or implicit disagreement.
EXAMPLES:- \{2 examples\}

- Score 2 — Moderate Severity (Explicit but Mild Conflict) Both statements explicitly refer to the same aspect, One gives light criticism, the other is mildly or significantly positive, Explicit but not extreme polarity.
EXAMPLES:- \{2 examples\}

- Score 3 — High Severity (Direct Strong Contradiction) Strongly worded positive vs. negative evaluation of the same aspect, Extremely polarized opposite judgments, Clear and fundamental extreme disagreement.
EXAMPLES:- \{2 examples\}

\textbf{REVIEW 1 CONTEXT}:
\{review\_1\}

\textbf{REVIEW 2 CONTEXT}:
\{review\_2\}

\textbf{EVIDENCE FROM REVIEW 1}:
\{s1\}

\textbf{EVIDENCE FROM REVIEW 2}:
\{s2\}

\textbf{DEBATE HISTORY}:
\{debate\_context\}

\textbf{YOUR ASSIGNED SCORE}: \{my\_score\}
\textbf{YOUR PREVIOUS REASONING}: \{my\_reasoning\}

\textbf{OPPONENT'S SCORE}: \{opponent\_score\}
\textbf{OPPONENT'S REASONING}: \{opponent\_reasoning\}

\textbf{CRITICAL INSTRUCTION}: You MUST defend your score of \{my\_score\}. You CANNOT change your score during the debate.

\textbf{RULES FOR EVIDENCE-BASED DEBATE}:
1. ONLY cite text that actually appears in the evidence or reviews above
2. Use direct quotes (in "quotation marks") when referencing the reviews
3. Point to specific words, phrases, or sentences that support your position
4. Counter your opponent by showing what they missed or misinterpreted in the actual text
5. Use simple language - avoid vague terms like "nuanced", "multifaceted", "complex interplay".
6. Don't use bold word or italics formatting in your response.
7. If opponent makes claims not supported by the text, point this out specifically

\textbf{Your task}:
1. Quote specific phrases from the evidence that support your score of \{my\_score\}
2. Identify flaws in opponent's reasoning by showing what the text actually says
3. Explain why your intensity level \{my\_score\} fits the criteria better than \{opponent\_score\}

Be assertive but fair. Focus on why YOUR intensity assessment is correct.

\textbf{Output format}:
\{\{
  "intensity": \{my\_score\},
  "reasoning": "your defense of score \{my\_score\} and counterarguments against opponent's score \{opponent\_score\}"
\}\}
    
\end{promptbox}

\begin{promptbox}{Adjudication Agent Prompt}

You are a judge evaluating a debate between two intensity scorers for peer-review contradictions.

\textbf{INTENSITY LEVELS}:
- Score 0 — No Contradiction (Compatible or Orthogonal Statements)
Statements refer to different aspects, topics, or evaluation criteria, OR
Statements discuss the same aspect but are fully compatible or consistent, OR
Any differences are descriptive, complementary, or additive rather than conflicting.

- Score 1 — Low Severity (Implicit Contradiction) One statement is generic, the other is specific, OR The conflict is indirect or interpretative, No strong positive/negative polarity, Weak or implicit disagreement.

- Score 2 — Moderate Severity (Explicit but Mild Conflict) Both statements explicitly refer to the same aspect, One gives light criticism, the other is mildly or significantly positive, Explicit but not extreme polarity.

- Score 3 — High Severity (Direct Strong Contradiction) Strongly worded positive vs. negative evaluation of the same aspect, Extremely polarized opposite judgments, Clear and fundamental extreme disagreement.

\textbf{TASK}:
1. Review the entire debate conversation
2. Examine the evidence from both reviews
3. Make a final judgment on the appropriate intensity score based on the reviews and the agents’ debate.
4. Provide clear reasoning for your decision.

\textbf{Consider}:
- Which arguments were most convincing?
- What does the evidence actually show?
- Does the contradiction meet the criteria for the claimed intensity level?
- Your decision must be based solely on the evidence and debate provided.

\textbf{REVIEW 1 CONTEXT}:
\{review\_1\}

\textbf{REVIEW 2 CONTEXT}:
\{review\_2\}

\textbf{EVIDENCE FROM REVIEW 1}:
\{s1\}

\textbf{EVIDENCE FROM REVIEW 2}:
\{s2\}

\textbf{COMPLETE DEBATE HISTORY}:
\{debate\_context\}

Based on the evidence and the debate, make your final judgment on the intensity score.

\textbf{Output format}:
\{\{
  "intensity": <0, 1, 2, or 3>,
  "reasoning": "your final judgment explaining why this is the correct intensity score"
\}\}
    
\end{promptbox}
\begin{promptbox}{TIDE (System Prompt)}
You are an expert at analyzing peer reviews of scientific papers and identifying contradictions between them and assign an intensity score to each contradiction. You are a contradiction extractor. Your task is not to generate contradictions, but to identify and extract ONLY the contradictions(ONLY if they exist) that are explicitly present between the two reviews provided.    
\end{promptbox}
\begin{promptbox}{TIDE (User Prompt)}
\textbf{TASK:}

You are given TWO peer reviews of the same scientific paper.
Your task is to identify where the two reviewers are contradicting. Therefore you need to EXTRACT the contradiction sentence pairs between the two reviews across multiple aspects and then assign an intensity score(1 to 3) to each contradiction evidence pair based on the severity scoring rubric provided below with a suitable reasoning for the assigned intensity score. 

\textbf{CONTRADICTION DEFINITION}:

A contradiction occurs when two statements make claims about the same aspect but opposing sentiment that cannot both be true simultaneously. Negations or disagreements alone do not constitute contradictions unless they pertain to the same factual claim. 

\textbf{ASPECTS TO CONSIDER}:

You must only consider contradictions that fall under one of the following aspects:
1. Substance
   (Insufficient experiments, weak analysis, missing ablations, lack of depth)
2. Motivation
   (Problem importance, relevance, impact, significance of the research)
3. Clarity
   (Writing quality, organization, readability, explanation of methods)
4. Meaningful Comparison
   (Fairness and completeness of comparisons to prior or baseline work)
5. Originality
   (Novelty of ideas, techniques, insights, or contributions)
6. Soundness
   (Correctness, validity, methodological justification, logical consistency)

\textbf{SEVERITY SCORING RUBRIC}:

- Score 1 — Low Severity (Implicit Contradiction)
  - One statement is generic, the other is specific
  - Conflict is indirect or interpretative
  - Weak or implicit disagreement
  - No strong positive or negative polarity

- Score 2 — Moderate Severity (Explicit but Mild Conflict)
  - Both statements refer to the same aspect
  - One is mildly critical, the other is moderately or strongly positive
  - Clear disagreement but not extreme

- Score 3 — High Severity (Direct Strong Contradiction)
  - Strongly positive vs strongly negative evaluation
  - Extremely polarized opposite judgments
  - Clear, fundamental, and direct conflict

\textbf{RULES}:
- Only use the listed aspects.
- Evidence must contain EXACTLY two sentences(one from review A and another one from review B) in the format:
  [sentence\_from\_Review\_1, sentence\_from\_Review\_2]
- The above two sentences must directly contradict each other on the SAME aspect. They must be claims that can't be true simultaneously.
- If multiple contradictions exist for the same aspect, include them all as separate entries.
- Use verbatim sentences from the reviews (no paraphrasing).
- Do NOT generate evidence that is not explicitly present in the reviews.(word by word evidence extraction)
- Do NOT include aspects with no contradictions.
- If there are no contradictions, return an empty array.
- Keep the scoring rubric open in mind while assigning intensity scores.

\textbf{REQUIRED OUTPUT FORMAT}:
Return a array where each element is an object with these fields:
- "evidence": An array with exactly two strings [Review\_A\_sentence, Review\_B\_sentence] OR "evidence": [] if no contradiction is found.
- "intensity\_reasoning": A detailed reasoning explaining why the assigned intensity score was chosen for the contradiction evidence pair.
- "intensity": An integer score (1 to 3) representing the severity of the contradiction based on the scoring rubric provided.

Following are two sample examples of contradictions from different review pairs(DONT COPY THESE EXAMPLES IN YOUR OUTPUT, THESE ARE ONLY FOR REFERENCE):

\textbf{Steps to follow}:
1. Read Both reviews carefully.
2. Identify contradictions between the two reviews following the contradiction definition.
3. Read the rules and validate the contradictions using the EXACT sentence pairs from both reviews while following the rules mentioned above.
4. Read the Scoring Rubric carefully.
5. Think and write the reasoning behind the intensity of contradiction according to the scoring rubric provided.
4. Assign an intensity score(1 to 3) to each contradiction evidence pair based on the reasoning you wrote and the severity scoring rubric provided.
6. Finally, return the contradictions in the REQUIRED OUTPUT FORMAT mentioned below or return empty array if no contradictions are found.

\textbf{Paper ID}: \{paper\_id\}

\textbf{INPUT REVIEWS}: 

Review A:
\{review\_a\}

Review B:
\{review\_b\}

The contradiction evidences(sentence pairs) between Review A and Review B are:
\end{promptbox}
\section{Case Studies} \label{Appendix: case_study}

\subsection{Case Study 1: DIA Agreement on Initial Severity} 
This example illustrates a scenario in which both DIA agents independently assign the same severity score to a contradiction without requiring further deliberation.
\begin{tcolorbox}[
  breakable,
  colback=gray!5,
  colframe=gray!60,
  title=\textbf{DIA Agreement on Initial Severity Score},
  fonttitle=\small\bfseries
]

\textbf{Evidence Statements:-}
\begin{quote}
\textit{Review 1:} The approach is novel and very interesting.
\end{quote}
\begin{quote}
\textit{Review 2:} All in all, the originality of the paper is lacking, the experimental setup is not convincing, and there are not much insights given by the paper into the novelty of method.
\end{quote}

\textbf{Aspect:} Originality

\vspace{0.5em}

\begin{tcolorbox}[
  breakable,
  colback=gray!4,
  colframe=gray!50,
  title=\textbf{Agent 1 (Score = 3)},
  fonttitle=\small\bfseries
]
The two pieces of evidence directly contradict each other on the aspect of the paper's originality. Review~1 states that the approach is novel and very interesting, while Review~2 claims that the originality of the paper is lacking. This constitutes a strong and direct contradiction, warranting a Score~3 (High Severity).
\end{tcolorbox}

\begin{tcolorbox}[
  breakable,
  colback=gray!4,
  colframe=gray!50,
  title=\textbf{Agent 2 (Score = 3)},
  fonttitle=\small\bfseries
]
The two statements present a direct and fundamental contradiction regarding the core merit of the paper. Review~1 explicitly affirms novelty, whereas Review~2 explicitly denies originality and insight. This diametrically opposed evaluation of the same aspect justifies a high-severity contradiction.
\end{tcolorbox}

\end{tcolorbox}
The two reviews make explicitly opposing claims about the paper’s originality: one affirms novelty, while the other directly denies it. Because both statements evaluate the same core aspect using strong and unambiguous language, the contradiction is immediately identifiable as high severity. As shown in the agent responses, both agents converge on a severity score of 3, demonstrating that clear, symmetric contradictions can be reliably detected and consistently scored by independent agents in the absence of ambiguity.

\subsection{Case Study 2: DIA Disagreement and the Role of Discussion}

This case illustrates how explicit multi-round discussion is necessary when agents disagree on the severity of a contradiction. Although both agents identify a clear conflict, they differ in interpreting the intensity of opposing evaluative language.

\begin{tcolorbox}[
  breakable,
  colback=gray!5,
  colframe=gray!60,
  title=\textbf{DIA Disagreement on Initial Severity Score},
  fonttitle=\small\bfseries
]

\textbf{Evidence Statements}
\begin{quote}
\textit{Review 1:} However, the ablation study and analysis on the model is weak and the authors only show experimental observations.
\end{quote}
\begin{quote}
\textit{Review 2:} The evaluation is thorough across the board, as far as I can tell.
\end{quote}

\textbf{Aspect:} Substance (Experimental Evaluation)

\vspace{0.5em}

\textbf{Round 1: Initial DIA Assessments}

\begin{tcolorbox}[
  breakable,
  colback=gray!4,
  colframe=gray!50,
  title=\textbf{Agent 1 (Score = 2)},
  fonttitle=\small\bfseries
]
The reviews explicitly disagree on the quality of the experimental evaluation. Review~1 describes the analysis as weak, while Review~2 describes it as thorough. The contradiction is clear, but the language is not maximally polarized, suggesting a moderate-severity conflict.
\end{tcolorbox}

\begin{tcolorbox}[
  breakable,
  colback=gray!4,
  colframe=gray!50,
  title=\textbf{Agent 2 (Score = 3)},
  fonttitle=\small\bfseries
]
The statements represent fundamentally opposing evaluations of the same aspect. Calling the analysis weak and observational directly conflicts with describing the evaluation as thorough across the board, indicating a high-severity contradiction.
\end{tcolorbox}

\vspace{0.5em}
\textbf{Rounds 2--4: Discussion and Reassessment}

\begin{tcolorbox}[
  breakable,
  colback=gray!3,
  colframe=gray!45,
  title=\textbf{Agent 1 (Maintains Score = 2)},
  fonttitle=\small\bfseries
]
Agent~1 argues that the wording in Review~1 criticizes depth rather than validity, and that Review~2’s assessment is hedged. These linguistic qualifiers suggest disagreement without maximal polarization, supporting a moderate-severity interpretation.
\end{tcolorbox}

\begin{tcolorbox}[
  breakable,
  colback=gray!3,
  colframe=gray!45,
  title=\textbf{Agent 2 (Maintains Score = 3)},
  fonttitle=\small\bfseries
]
Agent~2 emphasizes that phrases such as ``only show experimental observations'' and ``thorough across the board'' represent absolute and incompatible judgments, arguing that the contradiction reflects a fundamental evaluation conflict.
\end{tcolorbox}

\vspace{0.5em}
\textbf{Final Decision}

\begin{tcolorbox}[
  breakable,
  colback=gray!4,
  colframe=gray!60,
  title=\textbf{Judge (Final Score = 2)},
  fonttitle=\small\bfseries
]
Although the contradiction is explicit and concerns the same aspect, the negative assessment targets analytical depth rather than overall validity, and the positive assessment is subjectively hedged. The disagreement is therefore clear but not maximally polarized, leading to a final moderate-severity classification.
\end{tcolorbox}

\end{tcolorbox}

Through repeated discussion rounds, this case highlights how surface-level polarity can mask important differences in linguistic intensity. The deliberation process allows the framework to distinguish between strongly polarized contradictions and those where interpretive flexibility remains, resulting in a more calibrated severity judgment.

\subsection{Case Study 3: DIA Disagreement and the Role of Multi-Round Discussion}

This case study illustrates a scenario in which the DIA agents agree on the presence of a contradiction but disagree on its severity. Multiple discussion rounds are required to surface and resolve differences in how linguistic intensity and evaluative scope are interpreted.

\begin{tcolorbox}[
  breakable,
  colback=gray!5,
  colframe=gray!60,
  title=\textbf{DIA Disagreement on Initial Severity Score},
  fonttitle=\small\bfseries
]

\textbf{Evidence Statements}
\begin{quote}
\textit{Review 1:} However, the ablation study and analysis on the model is weak and authors only show experimental observations.
\end{quote}
\begin{quote}
\textit{Review 2:} The evaluation is thorough across the board, as far as I can tell.
\end{quote}

\textbf{Aspect:} Substance (Experimental Evaluation)

\vspace{0.5em}

\textbf{Round 1: Initial Severity Assessment}

\begin{tcolorbox}[
  breakable,
  colback=gray!4,
  colframe=gray!50,
  title=\textbf{Agent 1 (Score = 2)},
  fonttitle=\small\bfseries
]
The two pieces of evidence refer to the same aspect of the paper and express opposite opinions. Review~1 describes the ablation and analysis as weak, while Review~2 characterizes the evaluation as thorough. The contradiction is explicit, but the language is not extremely polarized, supporting a moderate-severity classification.
\end{tcolorbox}

\begin{tcolorbox}[
  breakable,
  colback=gray!4,
  colframe=gray!50,
  title=\textbf{Agent 2 (Score = 3)},
  fonttitle=\small\bfseries
]
Review~1 explicitly criticizes the ablation study and analysis as weak and merely observational, while Review~2 claims the evaluation is thorough across the board. These statements are interpreted as fundamentally incompatible evaluations of the same aspect, indicating a high-severity contradiction.
\end{tcolorbox}

What is gained in Round 1:

Both agents agree that the contradiction is explicit and concerns the same aspect. The disagreement is localized to the interpretation of severity rather than contradiction existence.

\vspace{0.5em}
\textbf{Round 2}

\begin{tcolorbox}[
  breakable,
  colback=gray!3,
  colframe=gray!45,
  title=\textbf{Agent 1 (Locked Score = 2)},
  fonttitle=\small\bfseries
]
The language used is not maximally severe. The term ``weak'' suggests insufficient analysis rather than a fundamental flaw, and ``thorough'' reflects a positive but not extreme assessment. Therefore, the contradiction remains of moderate severity.
\end{tcolorbox}

\begin{tcolorbox}[
  breakable,
  colback=gray!3,
  colframe=gray!45,
  title=\textbf{Agent 2 (Locked Score = 3)},
  fonttitle=\small\bfseries
]
The phrases ``only show experimental observations'' and ``across the board'' are interpreted as strong and absolute qualifiers. In the context of peer review, these represent extremely polarized judgments about evaluation quality.
\end{tcolorbox}

What is gained in Round 2:  

This round makes explicit the disagreement over whether the evaluative language should be interpreted as moderately critical or fundamentally dismissive.

\vspace{0.5em}
\textbf{Round 3}

\begin{tcolorbox}[
  breakable,
  colback=gray!3,
  colframe=gray!45,
  title=\textbf{Agent 1 (Locked Score = 2)},
  fonttitle=\small\bfseries
]
The criticism in Review~1 targets the depth of ablation and analysis rather than declaring the entire evaluation invalid. The contradiction is clear, but not maximally polarized.
\end{tcolorbox}

\begin{tcolorbox}[
  breakable,
  colback=gray!3,
  colframe=gray!45,
  title=\textbf{Agent 2 (Locked Score = 3)},
  fonttitle=\small\bfseries
]
Describing the analysis as weak and only observational undermines the scientific depth of the evaluation. When contrasted with ``thorough across the board,'' this represents a fundamental conflict in assessment.
\end{tcolorbox}

What is gained in Round 3:

The discussion clarifies whether the contradiction concerns partial inadequacy or a complete failure of the evaluation, which is central to severity classification.

\vspace{0.5em}
\textbf{Round 4}

\begin{tcolorbox}[
  breakable,
  colback=gray!3,
  colframe=gray!45,
  title=\textbf{Agent 1 (Locked Score = 2)},
  fonttitle=\small\bfseries
]
The positive claim is hedged by ``as far as I can tell,'' leaving room for missed weaknesses. Combined with the non-maximal negativity of ``weak,'' the contradiction remains moderately severe rather than extreme.
\end{tcolorbox}

\begin{tcolorbox}[
  breakable,
  colback=gray!3,
  colframe=gray!45,
  title=\textbf{Agent 2 (Locked Score = 3)},
  fonttitle=\small\bfseries
]
The absolutist phrasing ``across the board'' leaves no room for the weaknesses described in Review~1. This level of opposition is interpreted as maximally polarized.
\end{tcolorbox}

What is gained in Round 4:

This round consolidates all previously surfaced cues-linguistic strength, scope of criticism, and hedging-into a stable disagreement boundary.

\vspace{0.5em}
\textbf{Final Decision}

\begin{tcolorbox}[
  breakable,
  colback=gray!4,
  colframe=gray!60,
  title=\textbf{Judge (Final Score = 2)},
  fonttitle=\small\bfseries
]
Both reviews address the same aspect and clearly disagree, forming an explicit positive–negative contradiction. However, the negative wording critiques analytical depth rather than declaring the evaluation invalid, and the positive assessment is explicitly hedged by the phrase “as far as I can tell,” indicating limited confidence and leaving room for unobserved weaknesses. As a result, the polarity is clear but not extremely polarized, supporting a moderate-severity classification.
\end{tcolorbox}

\end{tcolorbox}

Across four discussion rounds, the framework progressively surfaces and evaluates linguistic intensity, evaluative scope, and hedging, with each round grounding the agents’ arguments more explicitly in the textual evidence. This iterative process moves the discussion from high-level explanations toward evidence-backed reasoning, enabling the judge to integrate all relevant signals and arrive at a calibrated final severity judgment that neither initial assessment alone could conclusively justify.

\end{document}